\documentclass[journal]{IEEEtran}

\usepackage{amsmath}
\usepackage{amsthm}
\usepackage{amsfonts}
\usepackage{amssymb}
\usepackage{epsfig}
\usepackage{psfrag}
\usepackage{graphicx}
\usepackage{float}
\usepackage{multirow}
\usepackage{tabularx}
\usepackage{slashbox}
\usepackage{color}
\usepackage{algorithm}
\usepackage{algorithmic}
\usepackage{lineno}
\usepackage{hyperref}
\usepackage{multicol}
\usepackage{commath}
\usepackage{stfloats}
\usepackage{wrapfig}
\usepackage{subcaption}
\usepackage{caption}
\usepackage{cleveref}
\usepackage{tikz}
\usetikzlibrary{shapes.geometric, arrows}

\begin{document}

\date{}

\title{A Closed-form Solution for Weight Optimization in Fully-connected Feed-forward Neural Networks}

\author{Slavisa~Tomic,
		João Pedro Matos-Carvalho,
        and Marko~Beko
\thanks{This work is funded by Fundação para a Ciência e Tecnologia under the project UIDB/50008/2020, UIDB/04111/2020, CEECINST/00147/2018/CP1498/CT0015 and 2021.04180.CEECIND, as well as Instituto Lus\'ofono de Investiga\c{c}\~ao e Desenvolvimento (ILIND) under Project COFAC/ILIND/COPELABS/1/2022.}
\thanks{Slavisa Tomic is with Center of Technology and Systems (UNINOVA-CTS) and Associated Lab of Intelligent Systems (LASI), 2829-516 Caparica, Portugal and COPELABS, Lusófona University, Campo Grande 376, 1749-024 Lisbon, Portugal (email: slavisa.tomic@ulusofona.pt).}
\thanks{João Pedro Matos-Carvalho is with Center of Technology and Systems (UNINOVA-CTS) and Associated Lab of Intelligent Systems (LASI), 2829-516 Caparica, Portugal and COPELABS, Lusófona University, Campo Grande 376, 1749-024 Lisbon, Portugal (email: joao.matos.carvalho@ulusofona.pt).}
\thanks{Marko Beko is with Center of Technology and Systems (UNINOVA-CTS) and Associated Lab of Intelligent Systems (LASI), 2829-516 Caparica, Portugal, Instituto de Telecomunicações, Instituto Superior Técnico, University of Lisbon, Lisboa, Portugal, and COPELABS, Lusófona University, Campo Grande 376, 1749-024 Lisbon, Portugal (email: marko.beko@tecnico.ulisboa.pt).}
}

\maketitle



\begin{abstract}
This work addresses weight optimization problem for fully-connected feed-forward neural networks. Unlike existing approaches that are based on back-propagation (BP) and chain rule gradient-based optimization (which implies iterative execution, potentially burdensome and time-consuming in some cases), the proposed approach offers the solution for weight optimization in closed-form by means of least squares (LS) methodology. In the case where the input-to-output mapping is injective, the new approach optimizes the weights in a back-propagating fashion in a single iteration by jointly optimizing a set of weights in each layer for each neuron. In the case where the input-to-output mapping is not injective (e.g., in classification problems), the proposed solution is easily adapted to obtain its final solution in a few iterations. An important advantage over the existing solutions is that these computations (for all neurons in a layer) are independent from each other; thus, they can be carried out in parallel to optimize all weights in a given layer simultaneously. Furthermore, its running time is deterministic in the sense that one can obtain the exact number of computations necessary to optimize the weights in all network layers (per iteration, in the case of non-injective mapping). Our simulation and experimental results show that the proposed scheme, BPLS, works well and is competitive with existing ones in terms of accuracy, but significantly surpasses them in terms of running time. To summarize, the new method is straightforward to implement, is competitive and computationally more efficient than the existing ones, and is well-tailored for parallel implementation.
\end{abstract}




\section{Introduction}
\label{sec:intro}

The advent of powerful computer hardware and workstations in the first decade of the twenty-first century enabled extraordinary ascent of machine learning. Since then, the area of machine learning (as a part of artificial intelligence) has been constantly evolving and attracting research interest, especially since it has been successfully applied to various academic or industrial problems~\cite{Abdou:2022}. Extremely powerful processing capabilities of such machines are allowing them to even transcend performance of human beings in some cases~\cite{Russell:2009}. Machine learning tools are particularly interesting for a wide range of practical problems in which it is difficult to establish accurate mathematical models~\cite{Abdou:2022}, since their solution often relies on solving some particular realizations of the problem with reasonably good accuracy~\cite{Nath:2014}. However, such problems could be solved more efficiently via machine learning techniques, where a set of sufficiently large samples (training data) are provided, based on which a machine could be able to \emph{learn} and \emph{make decisions} concerning a set of new samples (testing data).

Methods for solving convex programming problems with low convergence rates date back from the last decades of the 20th century, like~\cite{Nesterov:1982, MOM:1999}, for instance. More recently, several algorithms that successfully minimize regret (measured as the value of difference between a made decision and the optimal decision) in the online learning setting have been proposed~\cite{RMSProp:2013}-\cite{ADAM:2015}. All of the referred methods are related through the use of gradient descent, which is a relatively efficient optimization method given that the objective function is differentiable with respect to its parameters. In this case, the computation of first-order partial derivatives with respect to the parameters is of the same order of computational complexity as simply evaluating the function. The work in~\cite{Nesterov:1982} proposed a method that solves the problem in Hilbert space by constructing a minimizing sequence of points that are not relaxational. The latter property, together with a simple tuning of a parameter in the subdivision process to not start with $1$, allowed the authors to reduce the amount of computation at each step to a minimum and obtain an estimate of the convergence rate of their method. In~\cite{MOM:1999}, the author showed that introducing a momentum (MOM) term in gradient descent learning algorithms improves the speed of convergence and that it can increase the range of learning rate over which the system converges, by bringing certain eigen components of the system closer to critical damping. The work in~\cite{RMSProp:2013} showed how Long Short-term Memory recurrent neural networks could be used to generate complex sequences with long-range structure, by predicting a single data point at a time. The authors in~\cite{RMSProp:2013} used a form of stochastic gradient descent (SGD) where the gradients are divided by a running average of their recent magnitude and generated their parameter updates by using a momentum on the re-scaled gradient. In~\cite{AdaGrad:2011}, a family of sub-gradient methods that dynamically incorporate knowledge about certain properties of the data observed in previous iterations and perform more informative gradient-based learning was proposed. The authors generalized the dubious online learning paradigm to a sub-problem consisting of adapting an algorithm to fit a particular dataset by altering the proximal function and by automatically adjusting the learning rates for online learning. Over the recent years, works in~\cite{SGD:1998}-\cite{Deng:2013} showed that SGD method is an efficient optimization approach in various machine learning problems and that contributed to recent advances in deep learning. The authors in~\cite{NAG:2013} showed that applying SGD with a momentum with a well-designed random initialization and a particular type of slowly increasing schedule for the momentum parameter can train both deep and recurrent neural networks to levels of performance that were previously achievable only with Hessian-Free optimization. For this purpose, the authors employed a Nesterov's accelerated gradient (NAG). In~\cite{ADAM:2015}, an algorithm for first-order gradient-based optimization of stochastic objective functions, based on adaptive estimates of lower-order moments was presented. The method in~\cite{ADAM:2015} calculates adaptive learning rates individually for different parameters from estimates of first and second moments of the gradients by using initialization bias correction terms to avoid large step sizes that can lead to divergence. 

Even though the described solutions were derived for a much broader class of optimization problems, their application in weight optimization in neural networks is straightforward (and widely used nowadays) by exploiting them as optimizers that dynamically adjust the learning rate in a parameter estimation. In sharp contrast to the existing methods, this work proposes a novel approach for weight optimization whose solution is given in closed-form, by resorting to back-propagation (BP) and least squares (LS) methodology. The proposed solution optimizes the weights in a back-propagating (going from the output towards the input layer) manner by optimizing a set of weights in each layer for each neuron. The proposed scheme achieves weight optimization in a single iteration (per layer) and in just a few iterations for problems with injective and non-injective input-to-output mapping, respectively. What enhances even further the efficiency of the new approach is the fact that the computations in each layer are independent from each other, enabling them to be carried out in parallel. Moreover, it is worth mentioning that the running time of the proposed scheme is deterministic in the sense that one knows the exact number of computations necessary to optimize the weights in all network layers (per iteration, in the case of non-injective mapping) and is about $20$ times faster than the state-of-the-art solutions in the considered network architectures.



\section{The Proposed Approach}
\label{sec:solution}

\begin{figure}[!ht]
\centering
\hspace*{-0mm}\includegraphics[width=\linewidth]{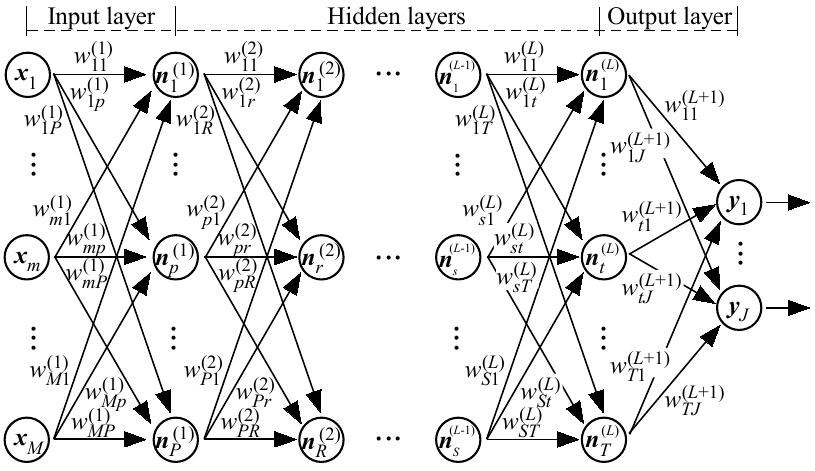}
\vspace*{-0mm}\caption{An example of a generic fully-connected neural network.}
\label{fig:fully_conn_netw}
\end{figure}

Consider a fully-connected feed-forward neural network\footnote{For the sake of simplicity of the following derivations, a network with no activation functions is considered as an illustrative example.} as illustrated in Fig.~\ref{fig:fully_conn_netw}. According to the network structure, the $j$-th output, $\boldsymbol{y}_j$, is obviously calculated as
\begin{equation}
\boldsymbol{y}_j = \displaystyle\sum_{j = 1}^J \displaystyle\sum_{t = 1}^T w_{tj}^{(L+1)}\boldsymbol{n}_{t}^{(L)},
\label{eq:outputs}
\end{equation}
for $j = 1, ..., J$, where the $i$-th neuron in the $l$-th layer (for $l \geq 2$) is given by
\begin{equation}
\boldsymbol{n}_{i}^{(l)} = \displaystyle\sum_{i = 1}^{N_l} \displaystyle\sum_{k = 1}^{N_{l-1}} w_{ki}^{(l)}\boldsymbol{n}_{k}^{(l-1)},
\label{eq:neuronsl2-lend}
\end{equation}
with $N_l$ and $N_{l-1}$ denoting the total number of neurons in the $l$-th layer and the $l-1$-th layer respectively (note that $N_0$ denotes the number of features of the input data, while $N_{L+1}$ denotes the number of outputs), while the neurons in the first layer ($l = 1$) are computed according to
\begin{equation}
\boldsymbol{n}_{p}^{(1)} = \displaystyle\sum_{p = 1}^P \displaystyle\sum_{m = 1}^M w_{mp}^{(1)}\boldsymbol{x}_{m}.
\label{eq:neuronsl1}
\end{equation}

To the best of our knowledge, until the present day, existing works focus their attention on training the neural networks by using back-propagation method in the following manner. The input data is multiplied with randomly generated weights in order to compute the output by exploiting the known network architecture. The so-produced output is then compared with the ground truth (i.e., the desired output) in order to update the weights by employing partial derivatives based on the chain rule as $$w_i^{(t+1)} = w_i^{(t)} + \eta \Delta w_i^{(t)},$$ where $w_i^{(t)}$ is the weight value that one is trying to optimize in the $t$-th iteration (epoch), $\eta$ is the learning rate (usually, either a fixed value or determined dynamically in each epoch by using an optimizer such as MOM~\cite{MOM:1999}, RMSProp~\cite{RMSProp:2013}, NAG~\cite{NAG:2013} or Adam~\cite{ADAM:2015}), and $\Delta w_i^{(t)}$ is the chain derivative of the output with respect to the $i$-th weight. Obviously, this method is iterative and non-deterministic in the sense that one cannot know how many iterations will the algorithm require to converge beforehand. Moreover, this approach is sub-optimal, meaning that it does not give any guarantees that the obtained solution is actually the optimal one, global-wise.

This work takes a different approach to accomplish weight optimization in a fully-connected feed-forward neural network. Given that the problem can be modeled as an injective function between the input and the output, the proposed solution requires a single iteration to optimize the weights of each neuron without resorting to derivatives and its solution is given in a closed-form (which allows for deterministic calculation of its running time). For problems with non-injective input-output relationship (such as classification problems), this work proposes a straightforward adaptation of the original idea that optimizes the weights in just a few \emph{epochs}, as described in the following text.

\begin{figure}[!ht]
\begin{subfigure}{.5\textwidth}
\hspace*{0mm}\includegraphics[width=\textwidth]{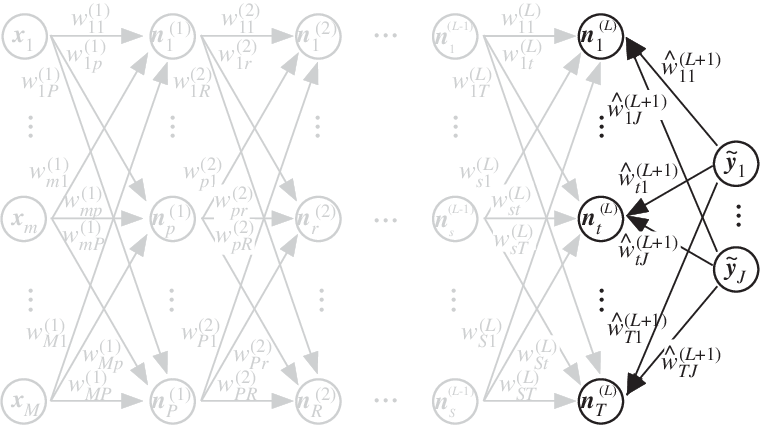}
\caption{Output (the last) layer}
\label{fig:trim_netw_out_lay}
\end{subfigure}
\vspace*{0mm}
\begin{subfigure}{.5\textwidth}
\hspace*{0mm}\includegraphics[width=\textwidth]{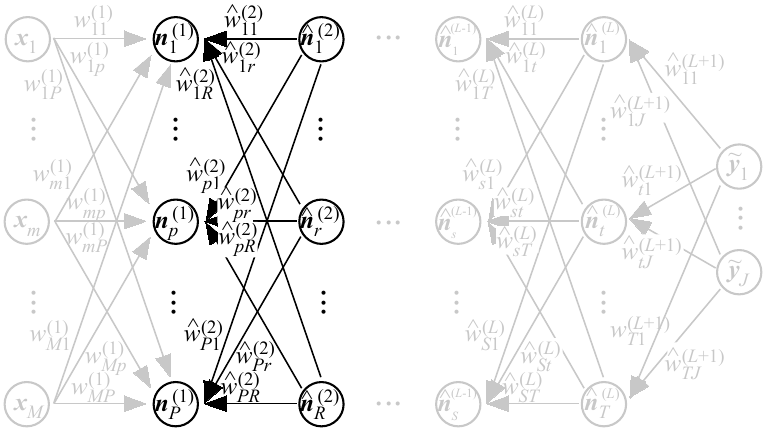}
\caption{Hidden (middle) layer}
\label{fig:trim_netw_hid_lay}
\end{subfigure}
\begin{subfigure}{.5\textwidth}
\hspace*{0mm}\includegraphics[width=\textwidth]{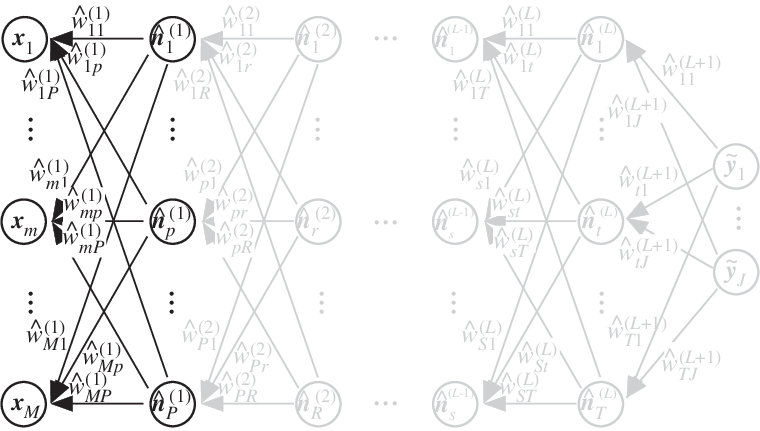}
\caption{Input (the first) layer}
\label{fig:trim_netw_in_lay}
\end{subfigure}
\vspace*{-0mm}
\caption{An example of the proposed network trimming for weight optimization of each neuron in a layer.}
\label{fig:trim_netw_all_lay}
\end{figure}

The proposed weight optimization process consists of optimizing the weights of each neuron individually in every layer in the back-propagating direction. First, similar to the existing approaches, randomly generated weights are assigned to the network in order to obtain the outputs, $\boldsymbol{y}_j$, according to~\eqref{eq:outputs}, by exploiting the known network architecture. Once $\boldsymbol{y}_j$s are available, the network is peeled layer by layer (in the opposite direction, from the output towards the input layers) and the corresponding weights are optimized based on an LS approach. More precisely, the weights of the $i$-th neuron ($i = 1, ..., N_l$) in the $l$-th layer, $\boldsymbol{w}_{i}^{(l)} = [w_{ki}^{(l)}]^T$, $k = 1, ..., N_{l-1}$, are optimized by solving the following LS problem
\begin{equation}
\hat{\boldsymbol{w}}_{i}^{(l)} = \underset{\boldsymbol{w}_{i}^{(l)}}{\text{arg\,min}} \, \|\boldsymbol{A}_l \boldsymbol{w}_{i}^{(l)} - \boldsymbol{b}_l\|^2.
\label{eq:LS_problem}
\end{equation}
The matrix $\boldsymbol{A}_l$ and the vector $\boldsymbol{b}_l$ in~\eqref{eq:LS_problem} are defined according to
\begin{equation}
\boldsymbol{A}_l =
\begin{cases}
\begin{bmatrix}
\vdots\\
\boldsymbol{x}_m\\
\vdots
\end{bmatrix}^T, & \text{if} \,\,\, l = 1,\\
\\
\begin{bmatrix}
\vdots\\
\boldsymbol{n}_{i}^{(l-1)}\\
\vdots
\end{bmatrix}^T, & \text{if} \,\,\, 1 < l \leq L + 1\\
\end{cases},
\nonumber
\end{equation}
\begin{equation}
\boldsymbol{b}_l =
\begin{cases}
[\hat{\boldsymbol{n}}_{i}^{(l)}]^T, & \text{if} \,\,\, 1 \leq l \leq L,\\
\\
[\widetilde{\boldsymbol{y}}_{l}]^T, & \text{if} \,\,\, l = L + 1
\end{cases},
\nonumber
\end{equation}
i.e., $\boldsymbol{A}_l \in \mathbb{R}^{d \times N_{l-1}}$ and $\boldsymbol{b}_l \in \mathbb{R}^{d\times 1}$, with $d$ representing the number of input data ($\boldsymbol{x}_m \in \mathbb{R}^{1\times d}$), $\widetilde{\boldsymbol{y}}_j$ denoting the desired output of the network (for a given input) and
\begin{equation}
\hat{\boldsymbol{n}}_{i}^{(l)} =
\begin{cases}
\frac{\widetilde{\boldsymbol{y}}_j - \displaystyle\sum_{\substack{t=1\\t\neq i}}^T \hat{w}_{tj}^{(l+1)} \boldsymbol{n}_{t}^{(l)}}{\hat{w}_{ij}^{(l+1)}}, & \text{if} \,\,\, l = L,\\
\\
\frac{\hat{\boldsymbol{n}}_{k}^{(l+1)} - \displaystyle\sum_{\substack{u=1\\u\neq i}}^{N_l} \hat{w}_{uk}^{(l+1)} \boldsymbol{n}_{u}^{(l)}}{\hat{w}_{ik}^{(l+1)}}, & \text{otherwise}
\end{cases},
\label{eq:neuron_upd}
\end{equation}
as illustrated in Figs.~\ref{fig:trim_netw_out_lay},~\ref{fig:trim_netw_hid_lay} and~\ref{fig:trim_netw_in_lay}.

Therefore, the solution of~\eqref{eq:LS_problem} is given in closed-form, i.e., the proposed weight optimization process boils down to just
\begin{equation}
\hat{\boldsymbol{w}}_{i}^{(l)} = \left( \boldsymbol{A}_l^T \boldsymbol{A}_l \right)^{-1} \left( \boldsymbol{A}_l^T \boldsymbol{b}_l \right).
\label{eq:solution}
\end{equation}

After this process has been performed for all layers of the network during the training phase, the network is left with the optimal weights (in the LS sense) for a given dataset.

To summarize, the proposed approach forces the network to produce the desired output by adjusting the weight values going into each neuron according to the proposed back-propagating LS (BPLS) methodology. The interesting thing is that the optimized weights in the last layer guarantee that the network will produce the desired output, but it also allows for updating the values of neurons across all layers. This is simply accomplished by exploiting the optimized weights and the desired output/updated value of the neurons in the preceding layer, together with the known network architecture in a back-propagating manner. In other words, each layer of the network is observed as a set of parallel systems with known inputs (the values computed by resorting to random weights) and known outputs (the desired ones). Hence, the optimization of weights is achieved in a simple and efficient manner in each layer and each neuron. The desired outputs in each layer are then used to push the information throughout the network by using the optimized weights to perform updates of the values of each neuron, after which the process is repeated for the succeeding layer.

It is worth mentioning that the proposed approach works also for networks with injective activation functions (such as sigmoid, hyperbolic tangent, exponential linear unit, softmax, softplus, softminus, etc.). In this case, the computation of the desired values in each neuron/output is adjusted to account for the use of the activation function, i.e., the updates of the neurons are obtained by calculating the inverse of the activation function whose argument is given on the right-hand side in~\eqref{eq:neuron_upd}. Finally, unlike the existing solutions, when the problem has an injective input-to-output mapping, the running time of the new scheme is deterministic in the sense that one can compute its exact running time, given that its total number of computations is exactly $N_l \times \left(L+1\right)$.

A pseudo-code of the proposed algorithm, called BPLS, is given in Algorithm~\ref{al:BPLS}. At line~$1$ of the algorithm, each of the weights in the network is randomly initialized according to a uniform random variable on the interval $[w_{\text{low}},w_{\text{upp}}]$. Line~$2$ is required so that all intermediate results (neuron values, $n_{il}, i = 1, ..., N_l, l = 1, ..., L$) for the random network are acquired. It is worth mentioning that, unlike the existing methods, lines~$5-10$ can be executed in parallel, i.e., one can obtain all weights for a given layer simultaneously by suitably rewriting the involved parameters in a matrix form.
\begin{algorithm}\footnotesize
\caption{~Pseudo-code of the Proposed BPLS Algorithm}
\begin{algorithmic}[1]
\REQUIRE $M:$ Number of inputs
\REQUIRE $\boldsymbol{x}_m:$ Input data $m = 1, ..., M$
\REQUIRE $L:$ Number of layers in the network
\REQUIRE $N_l:$ Number of neurons in the $l$-th layer, $l = 1, ..., L$
\REQUIRE $J:$ Number of outputs
\REQUIRE $\widetilde{\boldsymbol{y}}_j:$ Desired outcome at the $j$-th output
\STATE \textbf{Initialization:} Set $w_{ki}^{(l)}\sim\mathcal{U}[w_{\text{low}},w_{\text{upp}}], \,\, i = 1, ...,  N_l, \, k = 1, ...,  N_{l-1}, \, l = 1, ...,  L+1$
\STATE $\boldsymbol{y}_j \leftarrow$~\eqref{eq:outputs} {\fontfamily{qcr}\selectfont //Get output of random network}
\STATE $\boldsymbol{y}_j \leftarrow$~\eqref{eq:outputs} {\fontfamily{qcr}\selectfont //Get output of random network}
\FOR{$l = L+1 \, : \, -1 \, : \, 1$}
	\FOR{$i = 1 \, : \, 1 \, : \, N_{l-1}$}
		\STATE $\hat{\boldsymbol{w}}_i^{(l)} \leftarrow$~\eqref{eq:solution} {\fontfamily{qcr}\selectfont //Update weights in $l$-th layer}
		\IF{$l \neq 1$}		
			\STATE $\hat{\boldsymbol{n}}_{i}^{(l-1)} \leftarrow$~\eqref{eq:neuron_upd} {\fontfamily{qcr}\selectfont //Update neurons in $l-1$-th layer}
		\ENDIF
	\ENDFOR
\ENDFOR
\STATE \textbf{Return:} $\hat{\boldsymbol{y}}_j$ {\fontfamily{qcr}\selectfont //Use $\hat{\boldsymbol{w}}_i^{(l)}$ $\&$ netw. arch. for output at $j$}
\end{algorithmic}
\label{al:BPLS}
\end{algorithm}

The proposed BPLS solution can also be applied for problems that can be seen as non-injective systems, such as classification problems. In this case, the proposed algorithm is summarized in the pseudo-code in Algorithm~\ref{al:BPLS_class_prob}. Until line $14$, the main adjustment in our approach is that another LS problem is solved by using updated coefficient matrix of the system. This is required since it might not be possible to obtain accurate (enough) weight estimates from a high quantity of erroneous data. Once an initial estimate of the weights is acquired, the algorithm iteratively updates the weight values by exclusively using data for which an incorrect label has been given, rather than the entire dataset. This allows for faster execution of the algorithm compared to the existing ones and the procedure rewards previous weight estimates that reduce the number of missed classifications, line $20$.
\begin{algorithm}\footnotesize
\caption{~Pseudo-code for BPLS Adapted for Non-injective Systems}
\begin{algorithmic}[1]
\REQUIRE $M:$ Number of inputs
\REQUIRE $\boldsymbol{x}_m:$ Input data $m = 1, ..., M$
\REQUIRE $L:$ Number of layers in the network
\REQUIRE $N_l:$ Number of neurons in the $l$-th layer, $l = 1, ..., L$
\REQUIRE $J:$ Number of outputs
\REQUIRE $\widetilde{\boldsymbol{y}}_j:$ Desired outcome at the $j$-th output
\REQUIRE $\tau_{\text{max}}:$ Maximum number of iterations
\STATE \textbf{Initialization:} Set $w_{ki}^{(l)}\sim\mathcal{U}[w_{\text{low}},w_{\text{upp}}], \,\, i = 1, ...,  N_l, \, k = 1, ...,  N_{l-1}, \, l = 1, ...,  L+1$
\STATE $\boldsymbol{y}_j \leftarrow$~\eqref{eq:outputs} {\fontfamily{qcr}\selectfont //Get output of random network}
\FOR{$l = L+1 \, : \, -1 \, : \, 1$}
	\FOR{$i = 1 \, : \, 1 \, : \, N_{l-1}$}
		\IF{$l \neq 1$}
			\STATE $\widetilde{\boldsymbol{w}}_i^{(l)} \leftarrow$~\eqref{eq:solution} {\fontfamily{qcr}\selectfont //Initial update in $l$-th layer}
			\STATE $\hat{\boldsymbol{n}}_{i}^{(l-1)} \leftarrow$~\eqref{eq:neuron_upd} {\fontfamily{qcr}\selectfont //Update neurons in $l-1$-th layer}
			\STATE $\hat{\boldsymbol{A}}_l \leftarrow [\hdots \, (\hat{\boldsymbol{n}}_{i}{^{(l-1)}})^T \, \hdots]^T$ {\fontfamily{qcr}\selectfont //Upd. coeff. matrix}
			\STATE $\hat{\boldsymbol{w}}_i^{(l)} \leftarrow \left( \hat{\boldsymbol{A}}_l \hat{\boldsymbol{A}}_l^T \right)^{-1} \left( \hat{\boldsymbol{A}}_l \boldsymbol{b}_l^T \right)$ {\fontfamily{qcr}\selectfont //Estimate weights}
		\ELSE
			\STATE $\hat{\boldsymbol{w}}_i^{(l)} \leftarrow$~\eqref{eq:solution} {\fontfamily{qcr}\selectfont //Estimate weights in $1$-st layer}
		\ENDIF
	\ENDFOR
\ENDFOR
\STATE $\hat{\boldsymbol{y}}_j$ {\fontfamily{qcr}\selectfont //Estimate output at $j$}
\STATE $\mu^{(0)} \leftarrow (\hat{\boldsymbol{y}}_j == \widetilde{\boldsymbol{y}}_j)$ {\fontfamily{qcr}\selectfont //Estimated = desired labels?}
\STATE \textbf{Initialize:} $\tau \leftarrow 1$ and $\mu^{(\tau)} \leftarrow 10^6$
\WHILE{$\mu^{(\tau-1)} > \mu^{(\tau)}$ and $\tau \leq \tau_{\text{max}}$}
	\STATE  $\hat{\boldsymbol{w}}_{i,\text{miss}}^{(l)} \leftarrow$ repeat lines $3-14$ using inputs that produced a miss
	\STATE $\hat{\boldsymbol{w}}_i^{(l)} \leftarrow \left( 1 - \frac{\#\text{misses}}{\#\text{data}} \right) \hat{\boldsymbol{w}}_i^{(l)} + \frac{\#\text{misses}}{\#\text{data}} \hat{\boldsymbol{w}}_{i,\text{miss}}^{(l)}$
	\STATE $\hat{\boldsymbol{y}}_j$ {\fontfamily{qcr}\selectfont //Estimate output at $j$}
	\STATE $\mu^{(\tau)} \leftarrow (\hat{\boldsymbol{y}}_j == \widetilde{\boldsymbol{y}}_j)$ {\fontfamily{qcr}\selectfont //Check labels}
	\STATE $\tau \leftarrow \tau + 1$
\ENDWHILE
\STATE \textbf{Return:} $\hat{\boldsymbol{y}}_j$ {\fontfamily{qcr}\selectfont //Use $\hat{\boldsymbol{w}}_i^{(l)}$ $\&$ netw. arch. for output at $j$}
\end{algorithmic}
\label{al:BPLS_class_prob}
\end{algorithm}



\section{Discussion}
\label{sec:discuss}

In this section, an analysis of the proposed approach in terms of its computational complexity and feasibility of its solution is provided. The section is organized accordingly.

\subsection{Computational Complexity Analysis}
\label{subsec:complexity}

To analyse the computational complexity of the proposed approach, the Big O notation is adopted. Since the proposed approach allows for optimization of all weights in a given layer $l$ simultaneously,~\eqref{eq:solution} can be also be rewritten as
\begin{equation}
\hat{\boldsymbol{w}}^{(l)} = \left( \boldsymbol{A}_l^T \boldsymbol{A}_l \right)^{-1} \left( \boldsymbol{A}_l^T \boldsymbol{B}_l \right),
\label{eq:sol_parallel}
\end{equation}
where $\hat{\boldsymbol{w}}^{(l)} = [\hat{\boldsymbol{w}}_i^{(l)}] \in \mathbb{R}^{N_{l-1} \times N_{l}}$, $\boldsymbol{A}_l \in \mathbb{R}^{d \times N_{l-1}}$ and $\boldsymbol{B}_l = [\boldsymbol{b}_l] \in \mathbb{R}^{d \times N_{l}}$, with $d$ representing the number of input data used. Essentially, the proposed solution~\eqref{eq:sol_parallel} involves two matrix operations in the weight optimization process in each layer: multiplication and inversion. The former operation requires a complexity of $\mathcal{O}(\alpha \beta \gamma)$ for multiplication of an $\alpha \times \beta$ size matrix with a $\beta \times \gamma$ size matrix, while the latter operation requires a cubed complexity in the size of the matrix. Hence, by neglecting minor terms and focusing only on the dominant ones, the worst-case computational complexity of the proposed algorithm is $\mathcal{O}(h^3 + h^2 d)$, where $h = \max\{N_0, ..., N_{L}\}$.

\subsection{Feasibility Analysis}
\label{subsec:feasibility}

According to~\eqref{eq:sol_parallel}, let
\begin{equation}
\begin{array}{l}
f(\boldsymbol{w}^{(l)}) = \| \boldsymbol{A}_l \boldsymbol{w}^{(l)} - \boldsymbol{B}_l \|^2 = \\
= (\boldsymbol{w}^{(l)})^T \boldsymbol{A}_l^T \boldsymbol{A}_l \boldsymbol{w}^{(l)} - 2 (\boldsymbol{w}^{(l)})^T \boldsymbol{A}_l^T \boldsymbol{B}_l + \boldsymbol{B}_l^T \boldsymbol{B}_l
\end{array}
\nonumber
\end{equation}
denote the cost function to be minimized and $\mathcal{F}$ denote the set of feasible solutions. Since $\boldsymbol{A}_l^T \boldsymbol{A}_l$ is positive semi-definite by definition, the cost function is convex. Moreover, one can even guarantee strong convexity of the objective function by simply adding a (small) \emph{regularization} term, i.e.,
\begin{equation}
\begin{array}{l}
\tilde{f}(\boldsymbol{w}^{(l)}) = \| \boldsymbol{A}_l \boldsymbol{w}^{(l)} - \boldsymbol{B}_l \|^2 + \epsilon \|\boldsymbol{w}^{(l)}\|^2 = \\
(\boldsymbol{w}^{(l)})^T \left( \boldsymbol{A}_l^T \boldsymbol{A}_l + \epsilon \boldsymbol{I}_{N_{l-1}}  \right) \boldsymbol{w}^{(l)} - 2 (\boldsymbol{w}^{(l)})^T \boldsymbol{A}_l^T \boldsymbol{B}_l + \boldsymbol{B}_l^T \boldsymbol{B}_l,
\end{array}
\nonumber
\end{equation}
with $\epsilon = 10^{-6}$, for instance.

From the practical/engineering perspective, the two objective functions are virtually the same, but $\tilde{f}$ is strongly convex, meaning that it has a unique solution, since the matrix $\boldsymbol{A}_l^T \boldsymbol{A}_l + \epsilon \boldsymbol{I}_{N_{l-1}}$ is non-singular. Hence, the necessary and sufficient condition for $\boldsymbol{w}^{(l)}$ to be a minimizer of $\tilde{f}(\boldsymbol{w}^{(l)})$ is
\begin{equation}
\nabla \tilde{f}(\boldsymbol{w}^{(l)}) = 0 \Leftrightarrow \left( \boldsymbol{A}_l^T \boldsymbol{A}_l + \epsilon \boldsymbol{I}_{N_{l-1}}  \right) \boldsymbol{w}^{(l)} - \boldsymbol{A}_l^T \boldsymbol{B}_l = 0.
\nonumber
\end{equation}

Thus, it follows that $\mathcal{F} \neq \varnothing$, i.e., a (unique) solution for the proposed approach is feasible in practice.

%



\section{Numerical Results}
\label{sec:results}

This section validates the performance of the proposed approach, by comparing it to the following benchmark algorithms: AdaGrad in~\cite{AdaGrad:2011}, SGD in~\cite{SGD:1998}, NAG in~\cite{NAG:2013}, and Adam in~\cite{ADAM:2015}. The section first provides a set of numerical results for a couple of toy examples for which mathematical models between the input and the output are well-defined. Likewise, an experiment with real world datasets is performed by resorting to the MNIST and the Fashion-MNIST datasets for multi-class digit and fashion image recognition problems, respectively. The organization of this section is therefore done accordingly. It is worth mentioning that the initial weights for all neurons in all layers were drawn randomly from a uniform distribution on the interval $[-1, 1]$ in all considered scenarios. Moreover, for toy examples, all considered algorithms were executed in MATLAB by using CPU, whereas for experimental validation, all considered algorithms were executed in Python employing CUDA and PyTorch library with both CPU and GPU.
Lastly, the studied architectures were set empirically, but the generalization of all considered algorithms is straightforward to any type of architecture.

\subsection{Toy Examples}
\label{subsec:I/O_model}

\begin{figure}[!ht]
\centering
\hspace*{-0mm}\includegraphics[width=.85\linewidth]{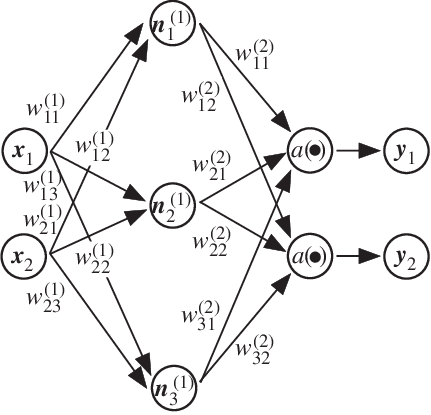}
\vspace*{0mm}\caption{The considered network with activation function in the toy example.}
\label{fig:toy_ex_netw}
\end{figure}

The performance of the proposed approach is validated here through a couple of functions that describe the known mathematical relationship between the input data and the desired output. Both linear and non-linear relationships are considered for a network comprising an input dataset and a bias fixed at value $1$ (resulting in $M=2$), with one hidden layer (i.e., $L=2$) composed of three neurons, $N_1 = 3$, and two distinct outputs, $J=2$, each one describing an independent function of the input data. In the latter case, sigmoid activation functions, defined as $a(x) = \frac{1}{1 + \exp \left\{ -x \right\} }$ for some $x \in \mathbb{R}$, are employed, as illustrated in Fig.~\ref{fig:toy_ex_netw}. For the training phase, the input dataset composed of a series of odd numbers ranging from $1$ to $9$, while the test is performed by a series of even numbers ranging from $2$ to $10$. Given the parsimonious quantity of data used, the maximum number of Epochs for the existing methods\footnote{All gradient descent-based algorithms were implemented by considering good default settings presented in~\cite{ADAM:2015}, including the learning rate of $\eta = 10^{-3}$, except for AdaGrad. In our experiments, we found that AdaGrad performs much better when considering $\eta = 10^{-1}$; hence, $\eta = 10^{-1}$ is adopted for AdaGrad in the toy examples.} was set to 1000. In all simulations presented in this section, the training dataset is corrupted by noise, $n_i$, modeled as a zero-mean Gaussian random variable with variance $\sigma^2$, i.e., $n_i \sim \mathcal{N}(0, \sigma^2)$. The principal metric used for assessing accuracy of an approach in this setting is the root mean squared error (RMSE), defined as $\text{RMSE} = \sqrt{\frac{1}{M_c}\sum_{m=1}^{M_c} \|\widetilde{\boldsymbol{y}}_{m} - \hat{\boldsymbol{y}}_{m}\|^2},$
where $\hat{\boldsymbol{y}}_{m}$ is the estimate of the desired output, $\widetilde{\boldsymbol{y}}_{m}$, in the $m$-th Monte Carlo, $M_c$, run.

\begin{figure}[!ht]
\begin{subfigure}{.5\textwidth}
\hspace*{0mm}\includegraphics[width=\textwidth]{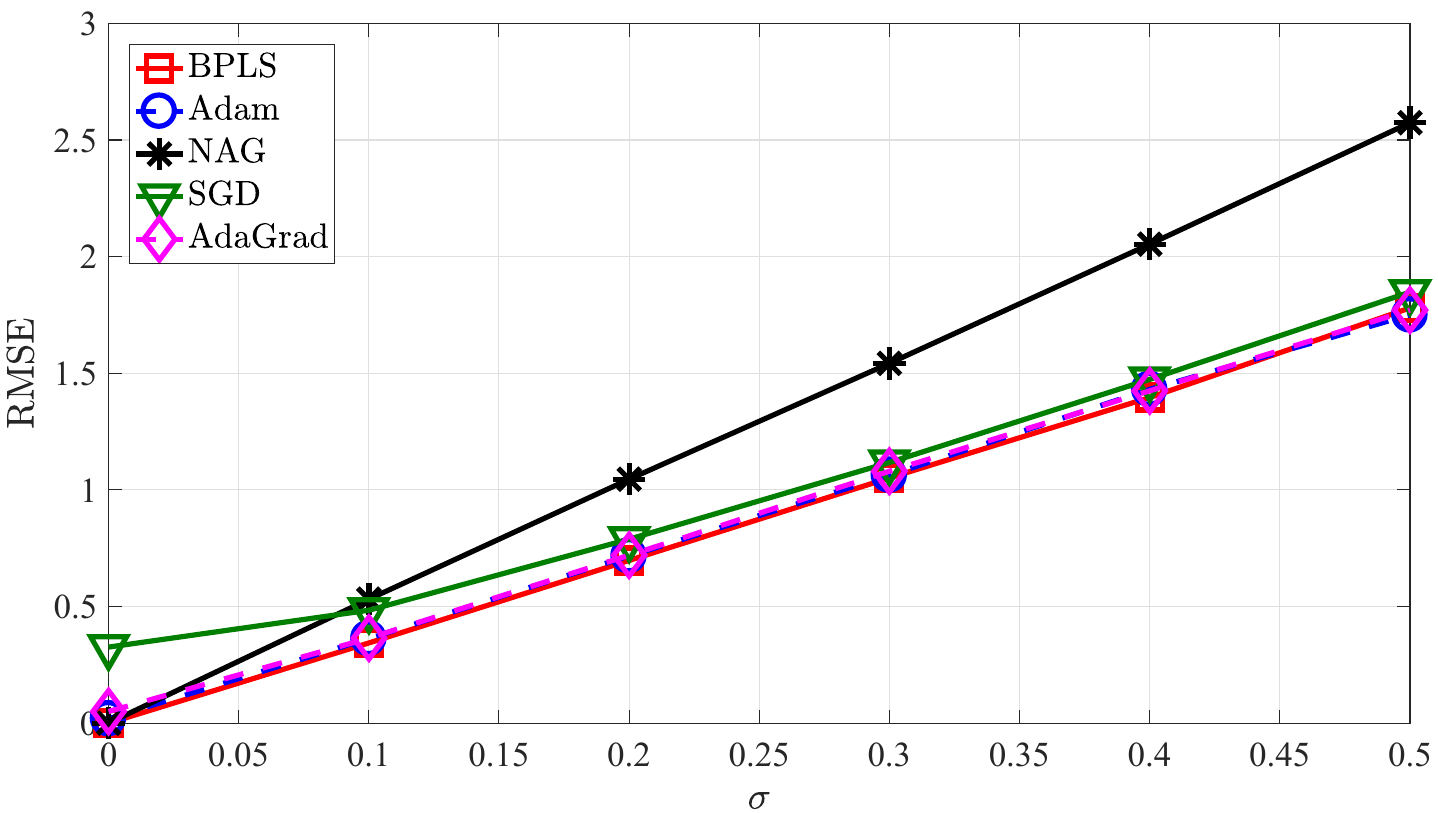}
\caption{Training phase}
\label{fig:I/O_lin_train}
\end{subfigure}
\vspace*{0mm}
\begin{subfigure}{.5\textwidth}
\hspace*{0mm}\includegraphics[width=\textwidth]{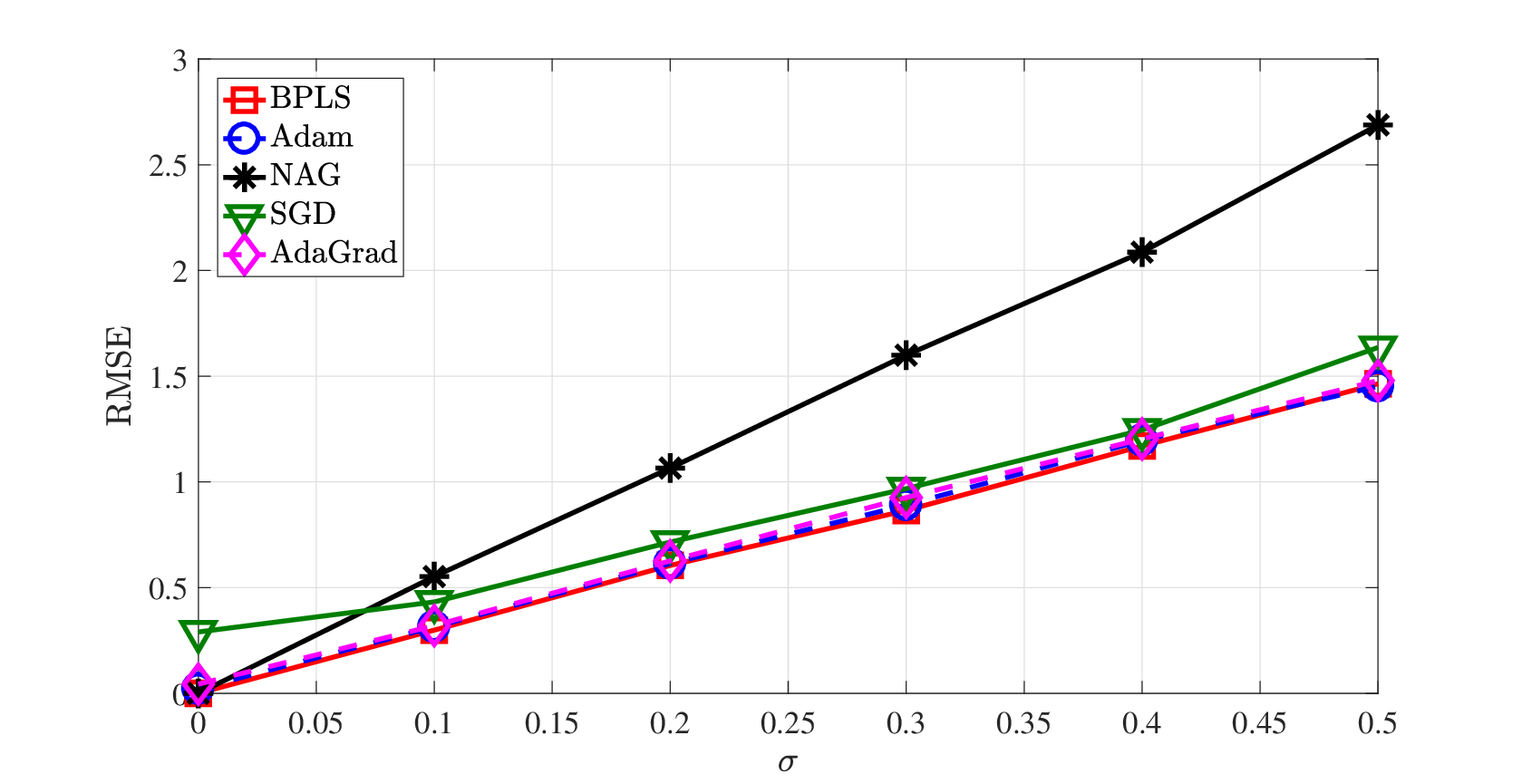}
\caption{Test phase}
\label{fig:I/O_lin_test}
\end{subfigure}
\vspace*{-0mm}
\caption{RMSE versus $\sigma$ illustration, for a well-defined linear mathematical model and $M_c = 1000$.}
\label{fig:I/O_lin}
\end{figure}

Fig.~\ref{fig:I/O_lin} illustrates RMSE versus $\sigma$ performance of the considered algorithms for a random linear input/output relationship, defined as $\widetilde{\boldsymbol{y}}_1 = -\frac{1}{3} \boldsymbol{x}_1 + 2$ and $\widetilde{\boldsymbol{y}}_2 = 2 \boldsymbol{x}_1 - 1$, for both training~\ref{fig:I/O_lin_train} and test~\ref{fig:I/O_lin_test} phases. As expected, all algorithms exhibit deterioration in their performance when the noise power grows. The proposed BPLS algorithm in Algorithm~\ref{al:BPLS} matches the performance of the state-of-the-art approach in the training phase and even slightly surpasses it in the test phase. This is an important accomplishment given that the proposed approach is significantly faster that the existing ones, and requires a single iteration (epoch) to obtain its final solution.

\begin{figure}[!ht]
\begin{subfigure}{.5\textwidth}
\begin{center}
\hspace*{0mm}\includegraphics[width=\textwidth]{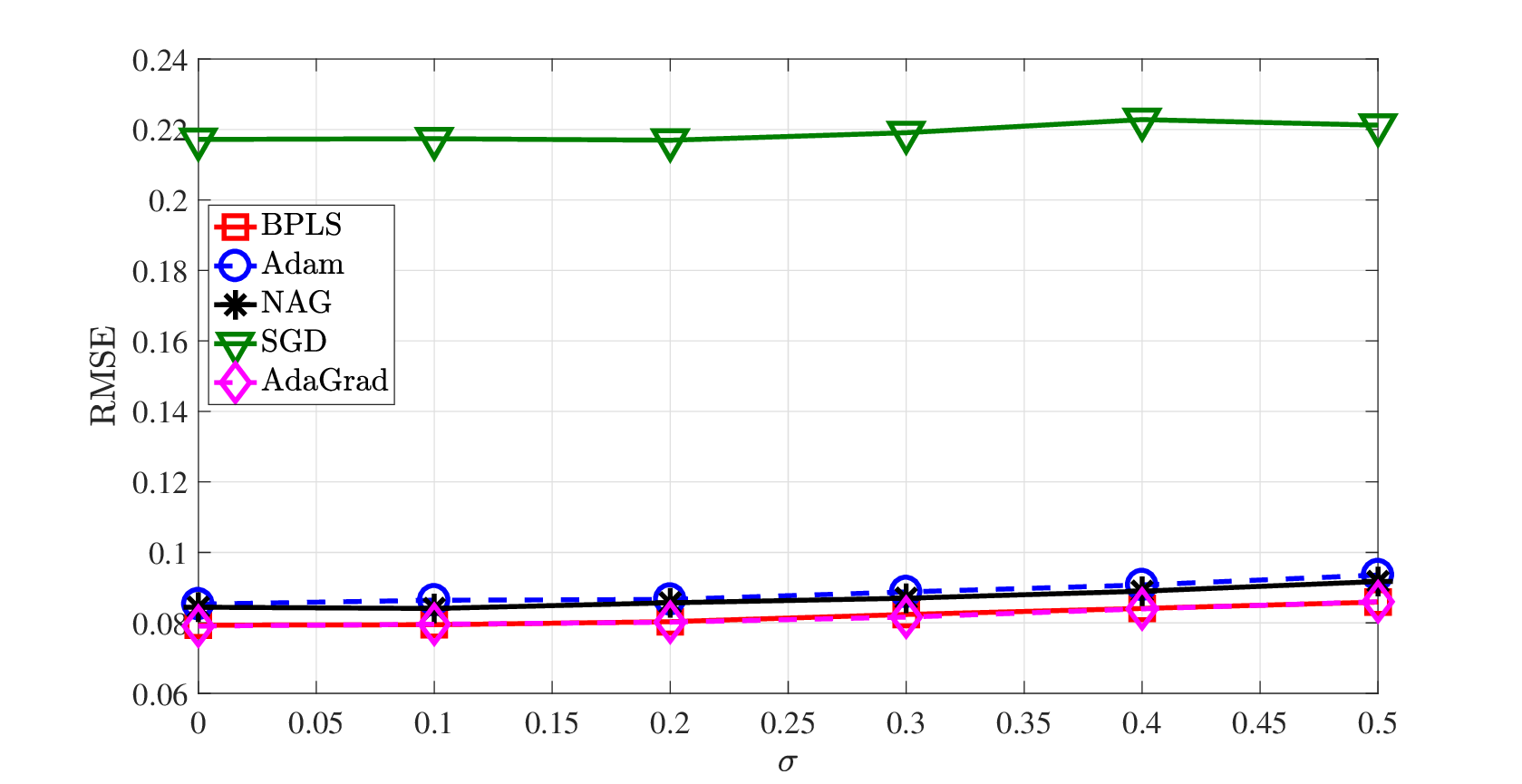}
\caption{Training phase}
\label{fig:I/O_nonlin_train}
\end{center}
\end{subfigure}
\vspace*{0mm}
\begin{subfigure}{.5\textwidth}
\begin{center}
\hspace*{0mm}\includegraphics[width=\textwidth]{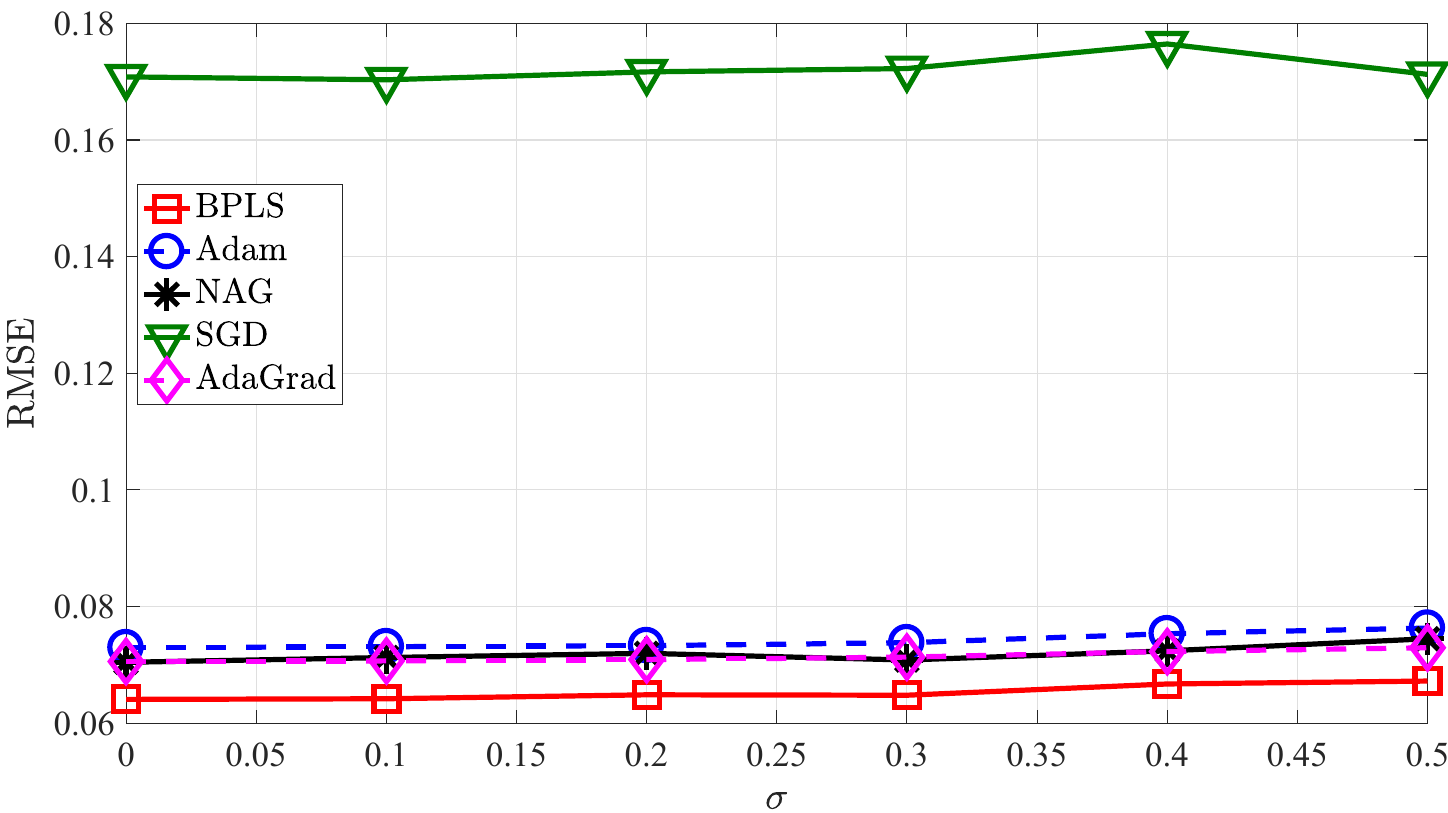}
\caption{Test phase}
\label{fig:I/O_nonlin_test}
\end{center}
\end{subfigure}
\vspace*{-0mm}
\caption{RMSE versus $\sigma$ illustration, for a well-defined non-linear mathematical model and $M_c = 1000$.}
\label{fig:I/O_nonlin}
\end{figure}

Fig.~\ref{fig:I/O_nonlin} illustrates RMSE versus $\sigma$ performance of the considered algorithms for a random non-linear input/output relationship, defined as $\widetilde{\boldsymbol{y}}_1 = \frac{1}{1+\exp\left\{ \log_{10} \left(\boldsymbol{x}_1^{-\frac{3}{2}}\right) \right\}}$ and $\widetilde{\boldsymbol{y}}_2 = \frac{1}{1+\exp\left\{ \boldsymbol{x}_1^{-\frac{1}{4}} \right\}}$, for both training~\ref{fig:I/O_nonlin_train} and test~\ref{fig:I/O_nonlin_test} phases. Similarly as in the previous case, the proposed approach virtually matches the best performance among the existing ones and outperforms them in the test phase somewhat more clearly. These results can be seen as an indicator that the existing methods slightly over-fitted the model in comparison with the proposed one.

\subsection{Real-world Experiments}
\label{subsec:MNIST}

\begin{figure}[!ht]
\begin{subfigure}{.49\textwidth}
\begin{center}
\hspace*{0mm}\includegraphics[width=.75\textwidth]{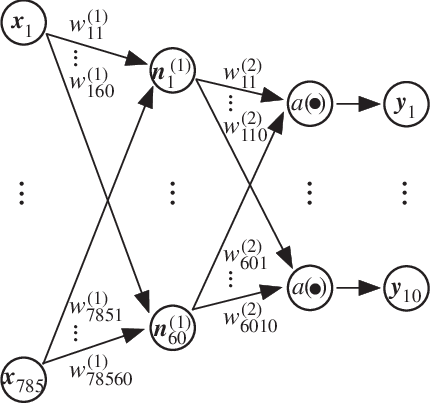}
\caption{Network with one hidden layer}
\label{fig:exp_ex_netw_1h}
\end{center}
\end{subfigure}
\vspace*{0mm}
\begin{subfigure}{.49\textwidth}
\begin{center}
\hspace*{0mm}\includegraphics[width=\textwidth]{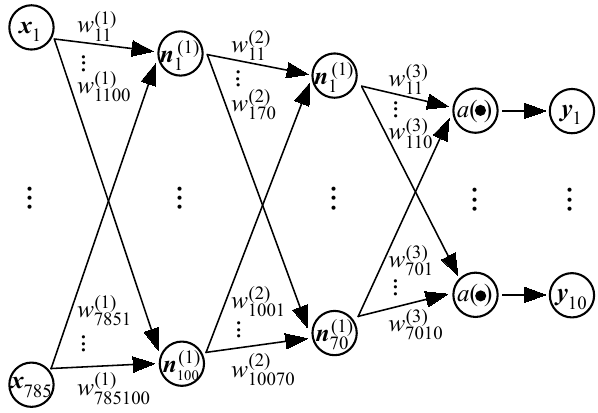}
\caption{Network with two hidden layers}
\label{fig:exp_ex_netw_2h}
\end{center}
\end{subfigure}
\vspace*{-0mm}
\caption{The considered network with softmax activation function in the experimental examples.}
\label{fig:exp_ex_netws}
\end{figure}

To experimentally assess the proposed method, two large, well-known and publicly-available image datasets (MNIST~\cite{MNIST,SGD:1998}, a dataset of handwritten Arabic digits, and Fashion-MNIST~\cite{Fashion-MNIST}, a dataset of fashion images) are used for training and testing the considered approaches. Both datasets contain a training set of $6 \times 10^4$ examples and a test set of $1 \times 10^4$ examples of $28 \times 28$ gray-scale images, associated with a label from 10 classes. Similar as before, the existing methods were implemented by considering the default settings presented in~\cite{ADAM:2015} with maximum $40$ Epochs and the learning rate $\eta = 10^{-3}$. Furthermore, for the existing algorithms, $\text{batch} = 1$ (in order to optimize the weight updates) and $M_c = 100$ was considered. In order to make the comparison completely fair, random weights were generated in each $M_c$ run and all methods were tested against exactly the same initial weights. For this setting, classification accuracy (CA), defined as the proportion of correctly labelled images to the total number of images, is chosen as the main performance metric. For both datasets, a fully-connected network with one hidden layer (i.e., $L = 2$) composed of $N_1 = 60$ and two hidden layers (i.e., $L = 3$) composed of $N_1 = 100$ and $N_2 = 70$ neurons, and $J = 10$ distinct outputs is considered, with softmax activation functions, defined as $a(\boldsymbol{x})_i = \frac{\exp\left\{x_i\right\}}{\sum_{j=1}^K \, \exp\left\{x_j\right\}}$ for $\boldsymbol{x} \in \mathbb{R}^{K}$, as illustrated in Fig.~\ref{fig:exp_ex_netws}.

\begin{figure}[!ht]
\begin{subfigure}{.5\textwidth}
\hspace*{0mm}\includegraphics[width=\textwidth]{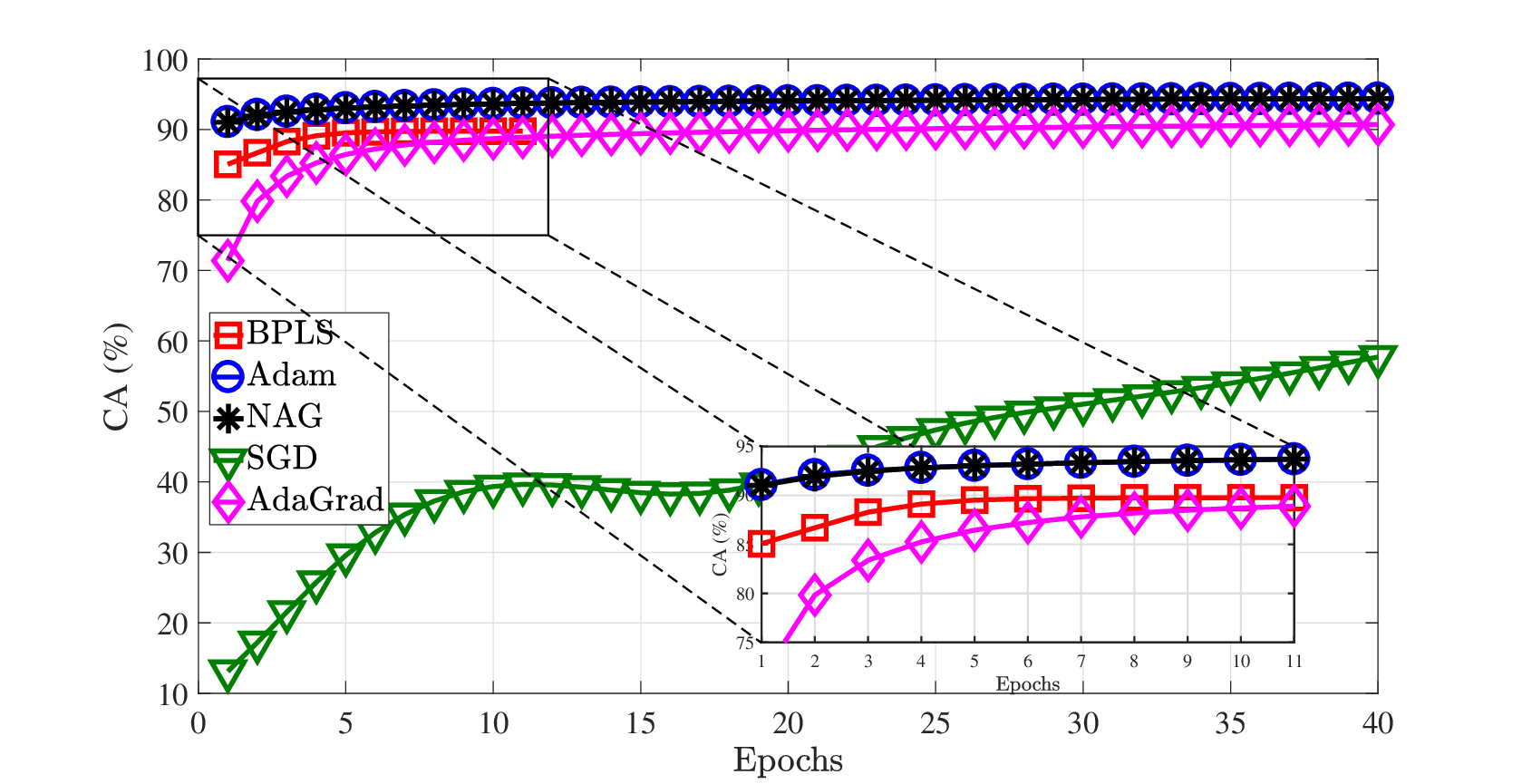}
\caption{Training phase}
\label{fig:train_MNIST}
\end{subfigure}
\vspace*{0mm}
\begin{subfigure}{.5\textwidth}
\hspace*{0mm}\includegraphics[width=\textwidth]{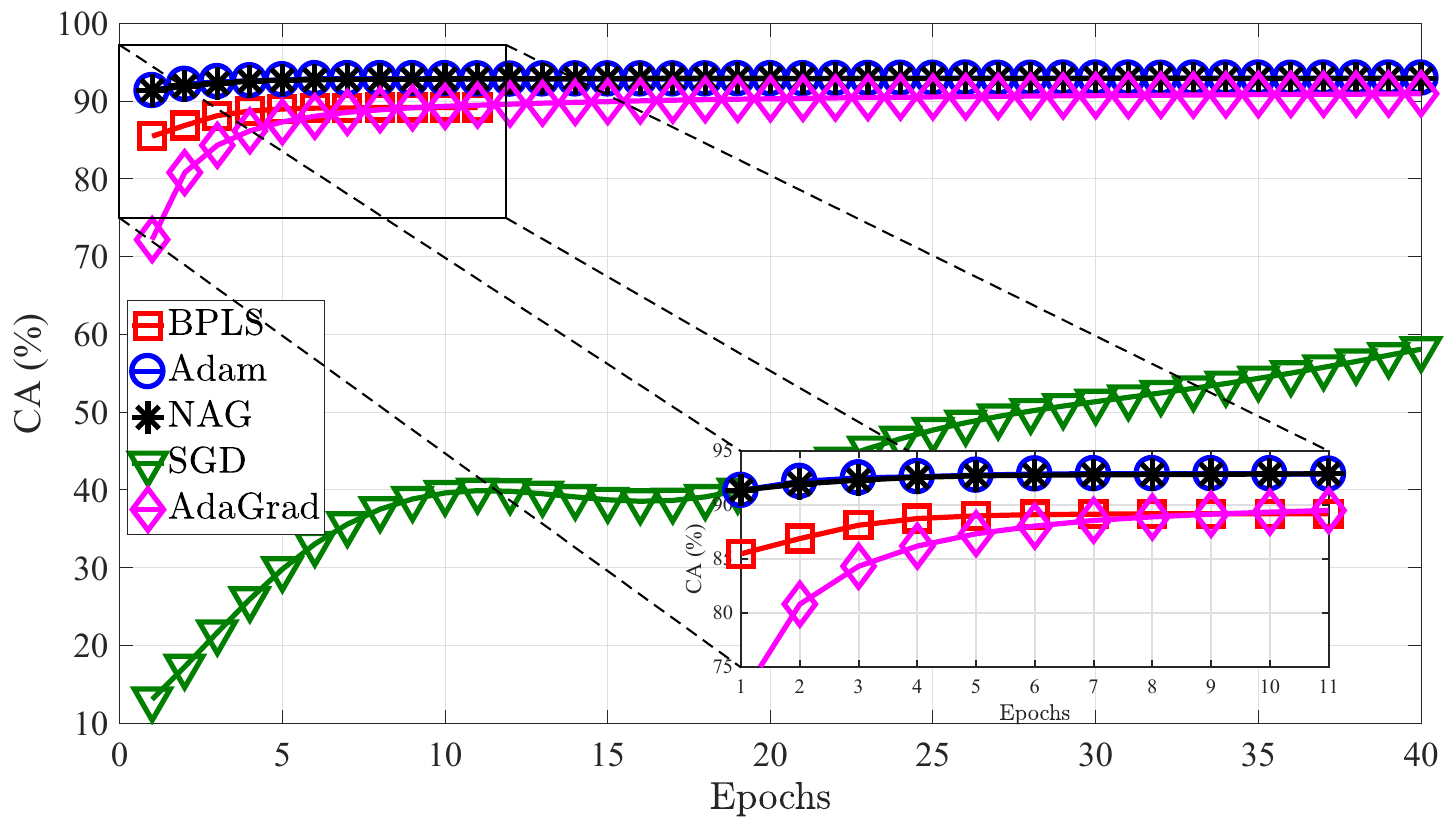}
\caption{Test phase}
\label{fig:test_MNIST}
\end{subfigure}
\vspace*{-0mm}
\caption{CA (\%) versus epochs illustration, with one hidden layer ($N_1 = 60$) using MNIST dataset.}
\label{fig:MNIST}
\end{figure}

Fig.~\ref{fig:MNIST} illustrates CA ($\%$) versus epochs performance of the considered algorithms for the MNIST training dataset in a fully-connected network with a single hidden layer, for both training~\ref{fig:train_MNIST} and test~\ref{fig:test_MNIST} phases. The figure shows that the proposed BPLS algorithm in Algorithm~\ref{al:BPLS_class_prob} requires only $11$ iterations in total to obtain its final solution, while the existing ones were allowed to make at most $40$ epochs (following the example set in~\cite{ADAM:2015}). One can observe that, apart from AdaGrad, the existing algorithms saturate within approximately $10$ epochs gaining very little-to-nothing in the remaining ones. The only method that does not converge within the considered maximum number of epochs is SGD, which indicates that it requires tuning of its learning rate or more epochs. From Fig.~\ref{fig:MNIST}, it can be seen that the proposed BPLS algorithms practically matches the performance of the state-of-the-art solutions, in both training and test phases.

\begin{figure}[!ht]
\begin{subfigure}{.5\textwidth}
\hspace*{0mm}\includegraphics[width=\textwidth]{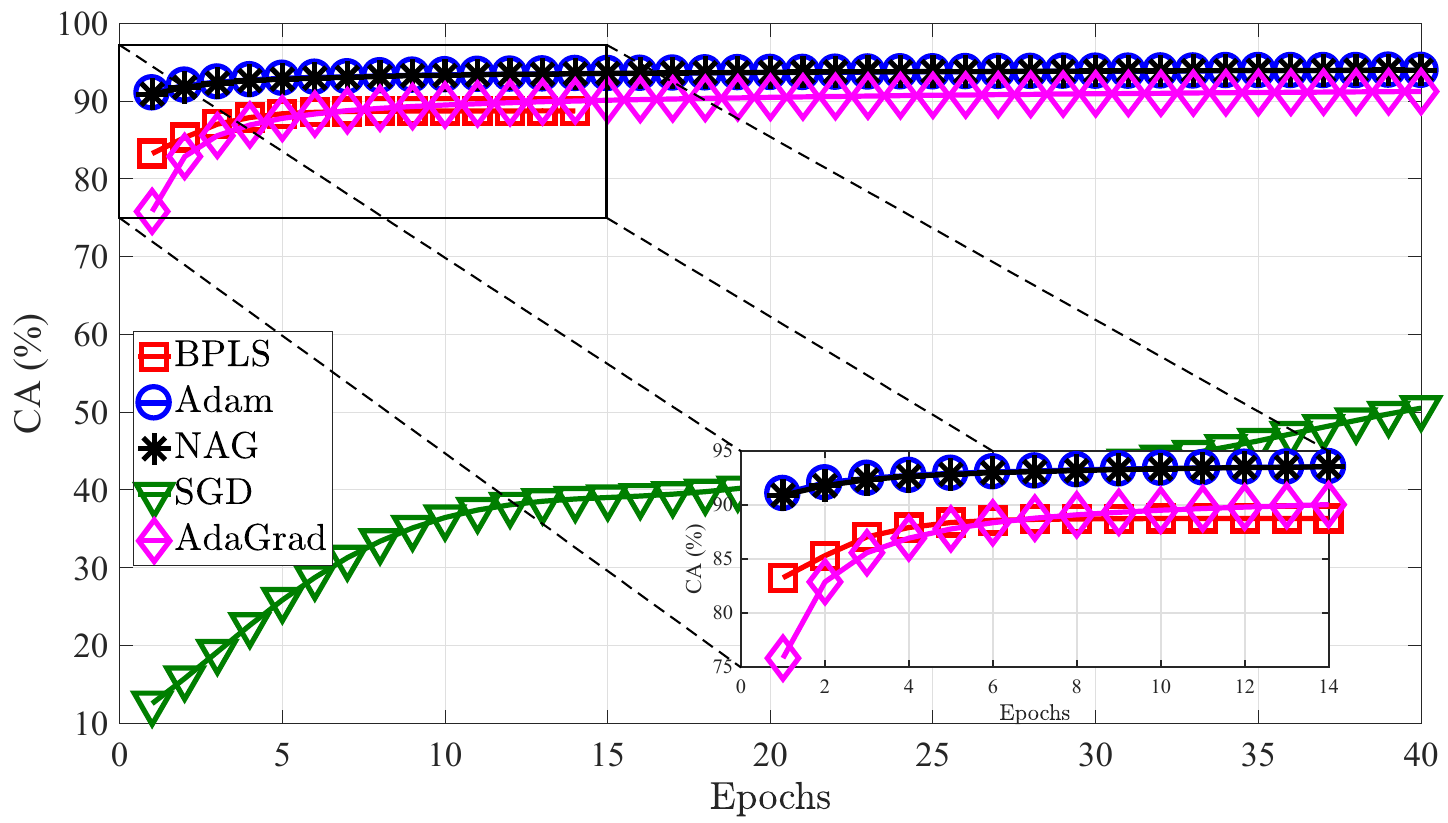}
\caption{Training phase}
\label{fig:train_MNIST_2h}
\end{subfigure}
\vspace*{0mm}
\begin{subfigure}{.5\textwidth}
\hspace*{0mm}\includegraphics[width=\textwidth]{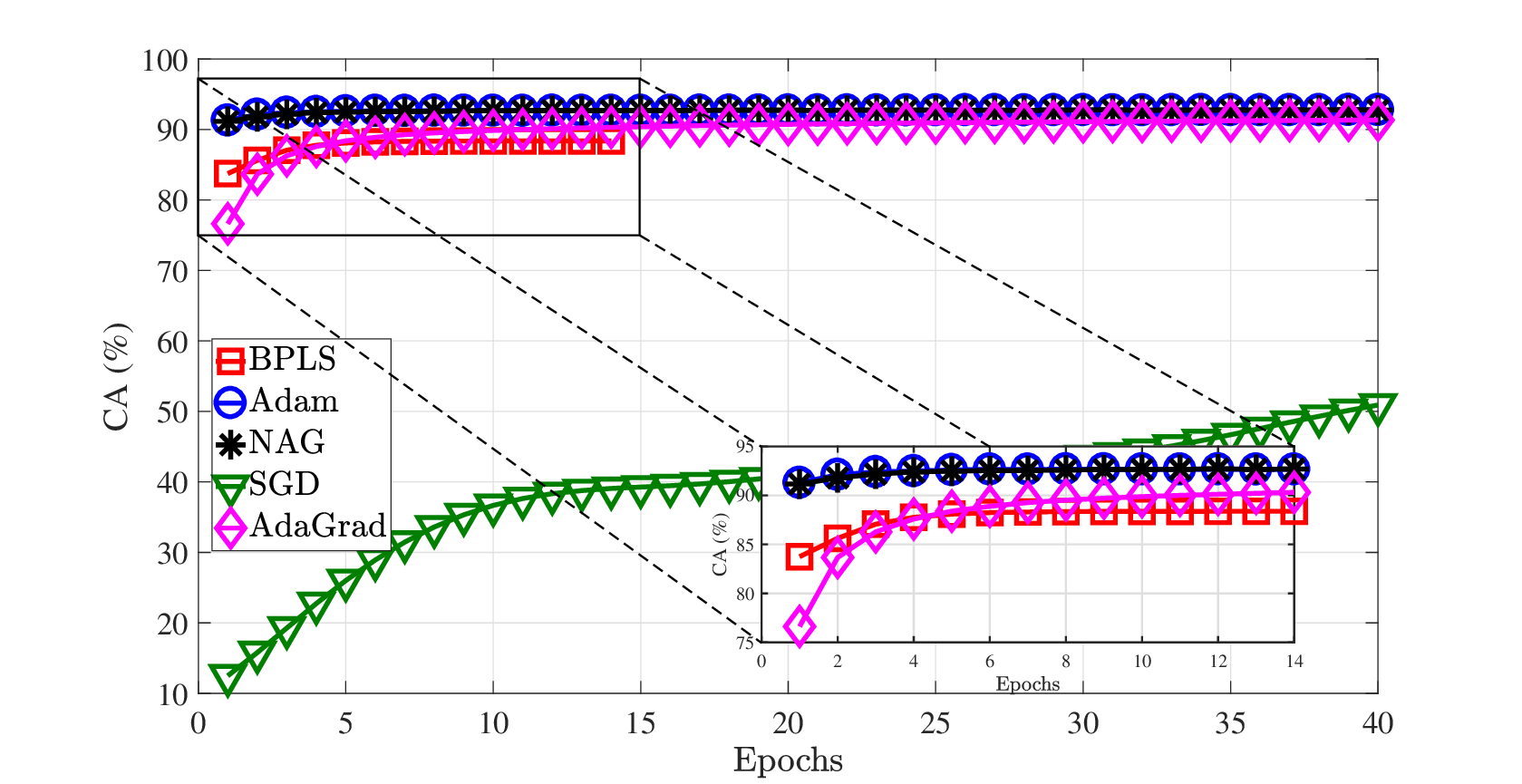}
\caption{Test phase}
\label{fig:test_MNIST_2h}
\end{subfigure}
\vspace*{-0mm}
\caption{CA (\%) versus epochs illustration, with two hidden layers ($N_1 = 100$ and $N_2 = 70$) using MNIST dataset.}
\label{fig:MNIST_2h}
\end{figure}

Fig.~\ref{fig:MNIST_2h} illustrates CA ($\%$) versus epochs performance of the considered algorithms for the MNIST training dataset in a fully-connected network with a pair of hidden layers, for both training~\ref{fig:train_MNIST_2h} and test~\ref{fig:test_MNIST_2h} phases. Naturally, in this setting, the proposed solution requires a somewhat higher number of iterations in total ($14$) than in the previous setting, although it gets saturated after roughly $5$, similar as most of the considered ones. Nevertheless, the performance of all considered methods is basically unchanged with the addition of an extra hidden layer, and the proposed BPLS is competitive with the best existing methods in both training and test phases.

\begin{figure}[!ht]
\begin{subfigure}{.5\textwidth}
\hspace*{0mm}\includegraphics[width=\textwidth]{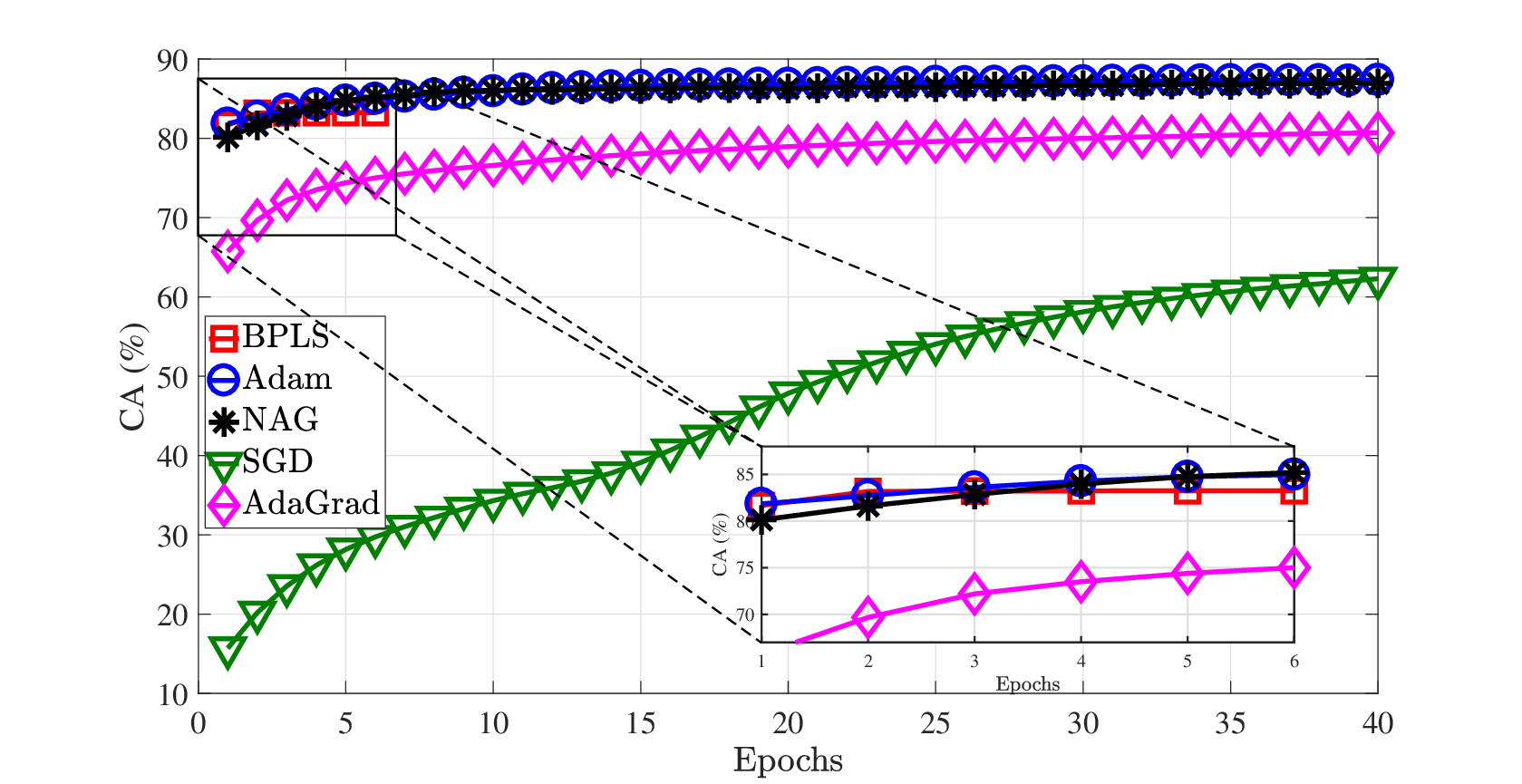}
\caption{Training phase}
\label{fig:train_FMNIST}
\end{subfigure}
\vspace*{0mm}
\begin{subfigure}{.5\textwidth}
\hspace*{0mm}\includegraphics[width=\textwidth]{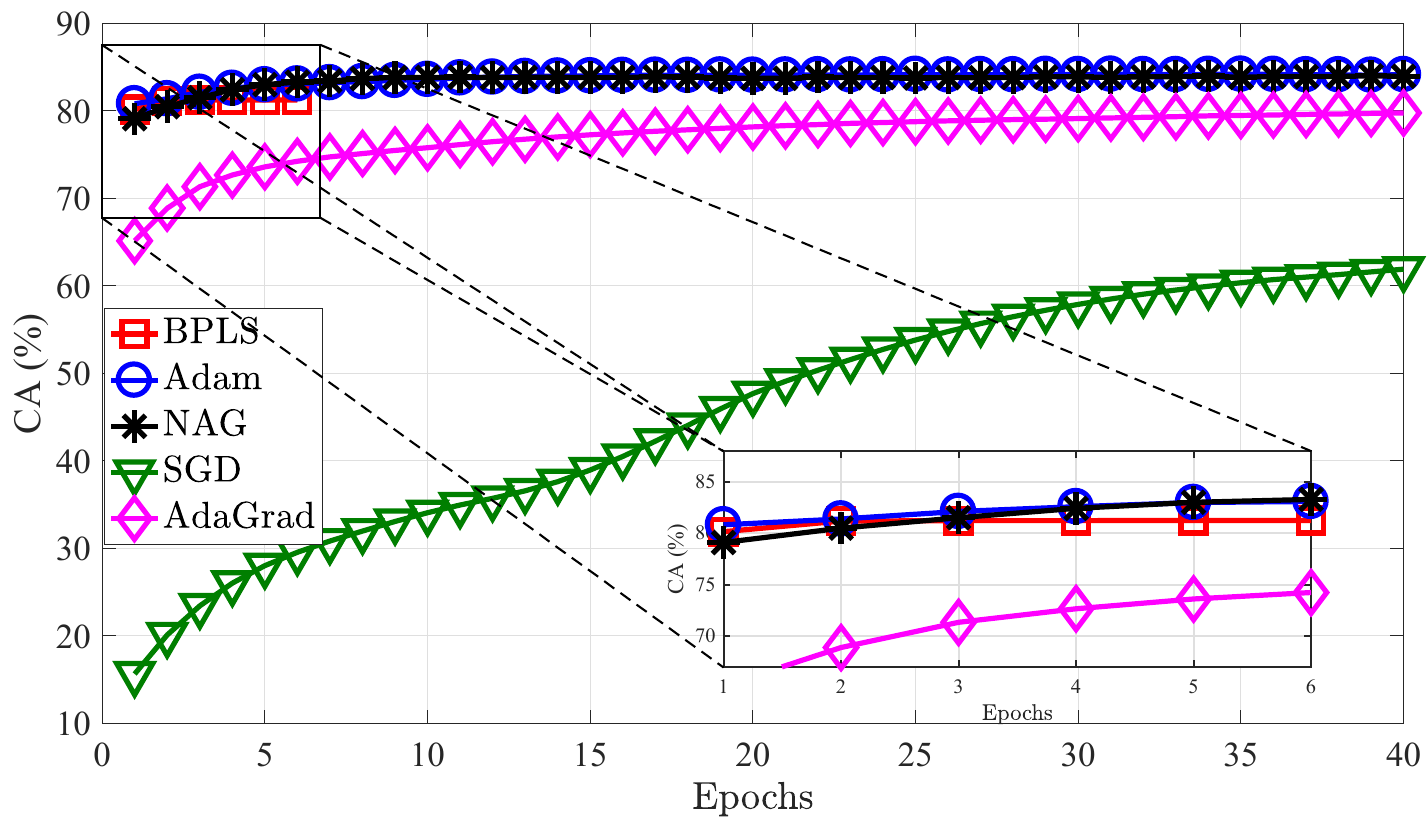}
\caption{Test phase}
\label{fig:test_FMNIST}
\end{subfigure}
\vspace*{-0mm}
\caption{CA (\%) versus epochs illustration, with one hidden layer ($N_1 = 60$) using Fashion-MNIST dataset.}
\label{fig:FMNIST}
\end{figure}

Fig.~\ref{fig:FMNIST} illustrates CA ($\%$) versus epochs performance of the considered algorithms for the Fashion-MNIST training dataset in a fully-connected network with a single hidden layer, for both training~\ref{fig:train_FMNIST} and test~\ref{fig:test_FMNIST} phases. Similar as for the MNIST dataset, Fig.~\ref{fig:FMNIST} shows that the proposed BPLS algorithm matches the performance of the existing methods and requires only $6$ iterations to obtain its final solution. Similarly, most of considered existing methods require the same number epochs to saturate and letting them to continue working leads to insignificant benefits.

\begin{figure}[!ht]
\begin{subfigure}{.5\textwidth}
\hspace*{0mm}\includegraphics[width=\textwidth]{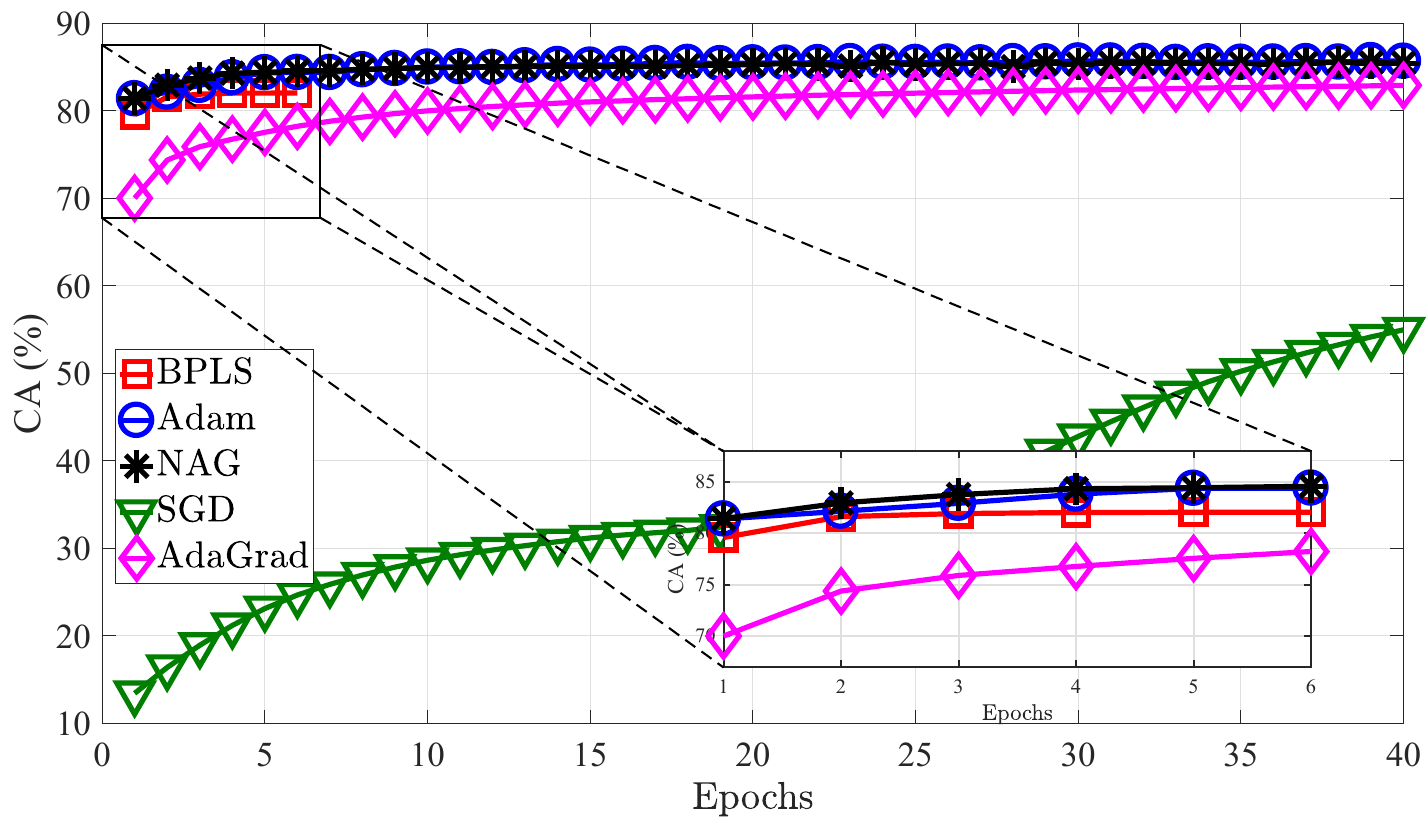}
\caption{Training phase}
\label{fig:train_FMNIST_2h}
\end{subfigure}
\vspace*{0mm}
\begin{subfigure}{.5\textwidth}
\hspace*{0mm}\includegraphics[width=\textwidth]{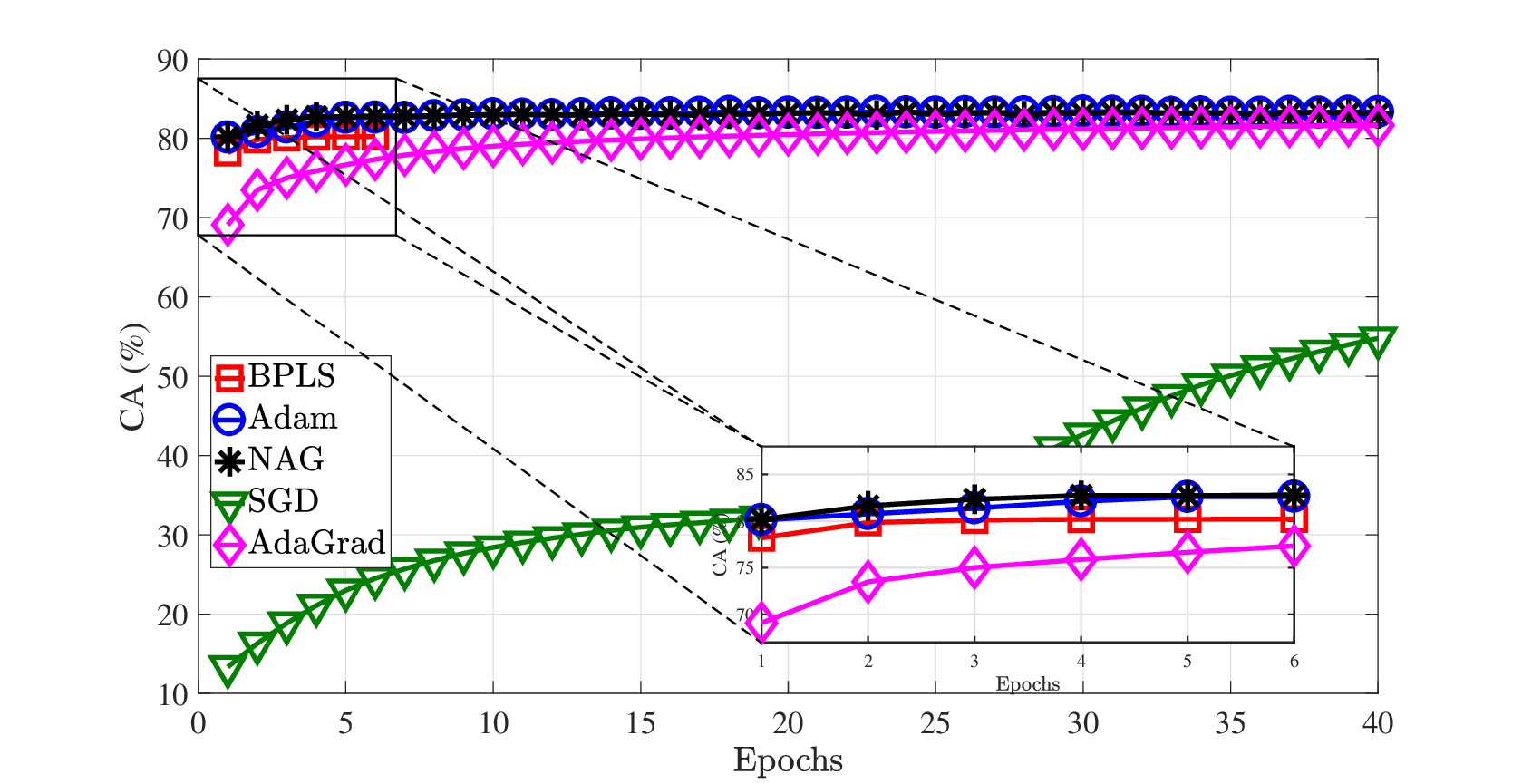}
\caption{Test phase}
\label{fig:test_FMNIST_2h}
\end{subfigure}
\vspace*{-0mm}
\caption{CA (\%) versus epochs illustration, with two hidden layers ($N_1 = 100$ and $N_2 = 70$) using Fashion-MNIST dataset.}
\label{fig:FMNIST_2h}
\end{figure}

Fig.~\ref{fig:FMNIST_2h} illustrates CA ($\%$) versus epochs performance of the considered algorithms for the Fashion-MNIST training dataset in a fully-connected network with a pair of hidden layers, for both training~\ref{fig:train_FMNIST_2h} and test~\ref{fig:test_FMNIST_2h} phases. Once again, the figure exhibits that introducing an additional hidden layer practically does not change the performance of the considered algorithms, and that the proposed BPLS is competitive with the best existing ones.

Table~\ref{tab:test_res} summarizes the best achieved CA (\%) results (on average) for the two datasets and two considered network architectures. The results presented in the table corroborate previous claim that the proposed solution practically matches the performance of the state-of-the-art methods in both training and test phases.
\begin{table*}[!ht]
\caption{The Best Achieved Training and Test Results for Different Image Datasets}
\vspace*{-2mm}
  \begin{center}
	\begin{tabular}{|c||*{9}{c|}*{9}{c|}*{9}{c|}*{9}{c|}*{9}{c|}}
	\hline
	\backslashbox{\textbf{Dataset}}{\textbf{Algorithm}} & \multicolumn{2}{c|}{\textbf{BPLS}} & \multicolumn{2}{c|}{\textbf{Adam}} & \multicolumn{2}{c|}{\textbf{NAG}} & \multicolumn{2}{c|}{\textbf{SGD}} & \multicolumn{2}{c|}{\textbf{AdaGrad}}\\ \hline
	\textbf{MNIST}, $L=2$ (training \& test) & 89.77 & 89.47 & 94.40 & 92.91 & 94.40 & 92.91 & 90.94 & 90.69 & 57.72 & 58.10\\ \hline
	\textbf{MNIST}, $L=3$ (training \& test) & 88.72 & 88.38 & 93.96 & 92.77 & 93.96 & 92.77 & 50.50 & 50.91 & 91.35 & 91.23\\ \hline
	\textbf{Fashion-MNIST}, $L=2$ (training \& test) & 83.25 & 81.23 & 87.37 & 83.94 & 86.93 & 83.94 & 62.30 & 61.89 & 80.71 & 79.75\\ \hline
	\textbf{Fashion-MNIST}, $L=3$ (training \& test) & 82.03 & 80.21 & 85.67 & 83.31 & 85.43 & 83.09 & 54.95 & 54.78 & 82.90 & 81.65\\ \hline
	\end{tabular}
   \end{center}
\label{tab:test_res}
\end{table*}

The average running times (in seconds) are exhibited in Table~\ref{tab:run_time}. It is worth mentioning that this analysis was performed on the machine with the following characteristics: CPU: Intel Core i7-8700 CPU \@ $3.20$GHz $\times$ $12$ with $64$GB RAM and GPU: NVIDIA GeForce RTX $2070$ SUPER/PCIe/SSE$2$ where the Fashion-MNIST dataset was employed in a fully-connected network with two hidden layers, where $\text{batch} = 1$ was considered for the existing algorithms. It is worth mentioning here that, by employing the existing PyTorch library, the state-of-the-art methods are fully optimized and adapted to CPU implementation. Even so, the table clearly shows the superiority of the proposed BPLS algorithm over the existing ones, obtaining its results about $20$ times faster.
\begin{table}[!ht]
\caption{Average Running Time (s) of the Considered Algorithms}
\vspace*{-2mm}
  \begin{center}
	\begin{tabular}{|c||c|c|c|c|c|}
	\hline
	\textbf{Algorithm} & \textbf{BPLS} & \textbf{Adam} & \textbf{NAG} & \textbf{SGD} & \textbf{AdaGrad}\\ \hline
	\textbf{Time (CPU)} & 0.26 & 4.93 & 4.94 & 4.83 & 4.86\\ \hline
	\textbf{Time (GPU)} & 0.17 & 4.85 & 4.82 & 4.69 & 4.73\\ \hline
	\end{tabular}
   \end{center}
\label{tab:run_time}
\end{table}

\begin{figure}[!ht]
\centering
\hspace*{-0mm}\includegraphics[width=.95\linewidth]{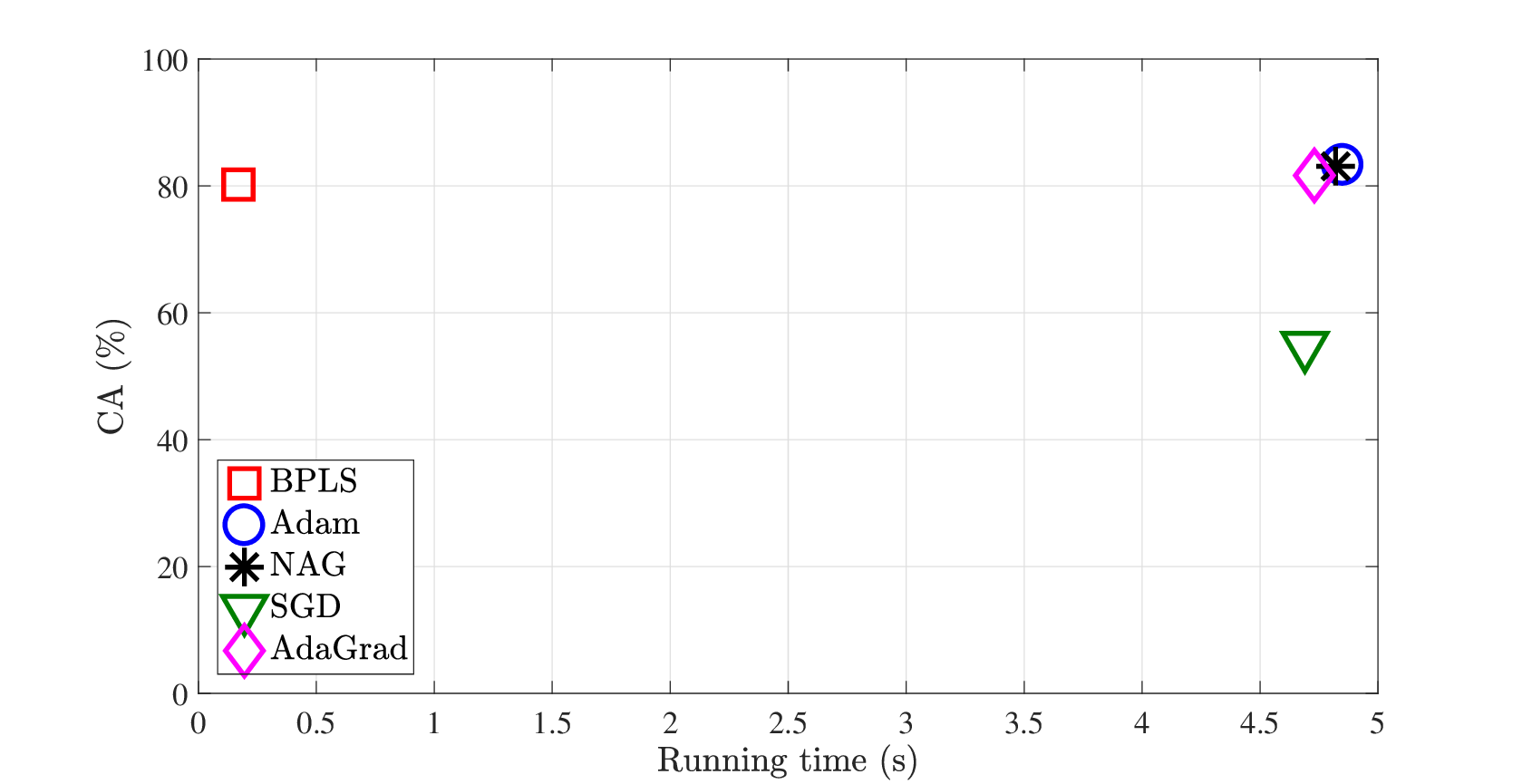}
\vspace*{0mm}\caption{CA (\%) versus average GPU running time (s) per epoch comparison for the Fashion-MNIST test dataset in a fully-connected network with two hidden layers.}
\label{fig:CA_vs_run_time_F_MNIST}
\end{figure}

To conclude the results analysis, Fig.~\ref{fig:CA_vs_run_time_F_MNIST} illustrates the CA (\%) versus average GPU running time (s) comparison for the Fashion-MNIST test dataset in a network with two hidden layers, which summarizes the main findings of this section. Ideally, one would desire the results to lie in the upper-left corner of the figure. It is exactly where the results of the proposed BPLS algorithm can be found, while the majority of results of the existing algorithms are situated in the right superior corner, illustrating the significant difference in running times.



\section{Conclusions}
\label{sec:conclusions}

This work proposed a novel approach to optimize weights for a fully-connected feed-forward neural network. Unlike the existing solutions that base their computation on iterative back-propagation and first-order gradient-based optimization schemes, the proposed solution optimizes the weights of each neuron in a single iteration and its solution is given in closed-form. It is founded on the least squares criterion, where a system of linear equations is derived by using back propagation according to the network architecture (going from right to left), where the solutions in the preceding layer are used to obtain the \emph{inputs} in the succeeding one. The new algorithm is suitable for parallel implementation and can thus optimize (in the LS sense) the weights of all neurons in a single layer simultaneously. The presented numerical experiments show that the proposed solution matches the performance of the state-of-the-art methods in terms of accuracy and surpasses them significantly in terms of running time (around $20$ times faster in the considered network architecture).


\begin{thebibliography}{99}


\bibitem{Abdou:2022} Mohamed A. Abdou, ``Literature Review: Efficient Deep Neural Networks Techniques for Medical Image Analysis'' \emph{Neural Computing and Applications,} vol. 34, pp.~5791--5812, February 2022.


\bibitem{Russell:2009} S. J. Russell and P. Norvig. \emph{Artificial Intelligence: A Modern Approach.} Prentice Hall, Upper Saddle River, NJ, USA, 2009.


\bibitem{Nath:2014} V. Nath , S. E. Levinson. \emph{Autonomous Robotics and Deep Learning.} Springer Cham, Switzerland, 2014.






\bibitem{Nesterov:1982} Y. E. Nesterov, ``A Method of Solving a Convex Programming Problem with Convergence Rate $\mathcal{O}\left(\frac{1}{k^2}\right)$,'' \emph{Doklady Akademii Nauk SSSR,} vol. 269, no. 3, pp. 543--547, July 1982.


\bibitem{MOM:1999} N. Qian, ``On the Momentum Term in Gradient Descent Learning Algorithms,'' \emph{Neural networks,} vol. 12, no. 1, pp. 145--151, January 1999.


\bibitem{RMSProp:2013} A. Graves, ``Generating Sequences with Recurrent Neural Networks,'' \emph{arXiv preprint arXiv:1308.0850,} 2013.


\bibitem{AdaGrad:2011} J. Duchi, E. Hazan, and Y. Singer, ``Adaptive Subgradient Methods for Online Learning and Stochastic
Optimization,'' \emph{The Journal of Machine Learning Research,} vol. 12, pp. 2121--2159, July, 2011.


\bibitem{SGD:1998} Y. LeCun, L. Bottou, Y. Bengio, and P. Haffner, ``Gradient-based Learning Applied to Document Recognition,'' \emph{Proceedings of the IEEE}, vol. 86, no. 11 pp. 2278--2324, November 1998.


\bibitem{Hinton:2012} G. Hinton, L. Deng, D. Yu, et. al. ``Deep Neural Networks for Acoustic Modeling in Speech Recognition: The Shared Views of Four Research Groups,'' \emph{IEEE Signal Processing Magazine,} vol. 29, no. 6, pp. 82--97, November 2012.


\bibitem{Krizhevsky:2012} A. Krizhevsky, I. Sutskever, G. E. Hinton, ``ImageNet Classification with Deep Convolutional Neural Networks,'' \emph{Communications of the ACM,} vol. 60, no. 6 pp. 84--90, May 2017.


\bibitem{Deng:2013} L. Deng, J. Li, J. T. Huang, et al. ``Recent Advances in Deep Learning for Speech Research at Microsoft,'' \emph{The 38th International Conference on Acoustics, Speech, and Signal Processing,} Vancouver, BC, Canada, pp. 8604--8608, May 26--31, 2013.


\bibitem{NAG:2013} I. Sutskever, J. Martens, G. Dahl, and G. Hinton, ``On the Importance of Initialization and Momentum in Deep Learning,'' \emph{The 30th International Conference on Machine Learning,} Atlanta, GA, USA, pp. 1139--1147, June 16--21, 2013.


\bibitem{ADAM:2015} D. P. Kingma and J. Ba, ``Adam: A Method for Stochastic Optimization,'' \emph{3rd International Conference for Learning Representations,} San Diego, CA, USA, pp. 1--15, May 7--9, 2015.




\bibitem{MNIST} Y. LeCun, C. Cortes, and C. J. C. Burges, ``The MNIST Database of Handwritten Digits,'', 1999. Retrieved from \url{http://yann.lecun.com/exdb/mnist/}


\bibitem{Fashion-MNIST} H. Xiao, K. Rasul, and R. Vollgraf, ''Fashion-MNIST: A Novel Image Dataset for Benchmarking Machine Learning Algorithms,'' \emph{arXiv preprint arXiv:1708.07747}, August, 2017.


\end{thebibliography}
\end{document}